\pdfoutput=1

\documentclass{tlp}
\usepackage{aopmath}

\usepackage{epsfig}

\usepackage[english]{babel}

\newtheorem{thm}{Theorem}

\newcommand{\answerset}[1]{{\it answerset}(#1)}
\renewcommand{\ttdefault}{cmtt}

\newtheorem{definition}{Definition} 
\newtheorem{example}{Example} 

\begin{document}
\bibliographystyle{acmtrans}

\long\def\comment#1{}

\title[Building Rules on Top of Ontologies for the Semantic Web with ILP]{Building Rules on Top of Ontologies\\for the Semantic Web\\with Inductive Logic Programming}

\author[F.A. Lisi]
{FRANCESCA A. LISI \\
Dipartimento di Informatica \\
Universit\`{a} degli Studi di Bari \\
Via Orabona 4\\
70125 Bari, Italy\\
E-mail: lisi@di.uniba.it
}

\pagerange{\pageref{firstpage}--\pageref{lastpage}}
\volume{\textbf{10} (3):}
\jdate{September 2007}
\setcounter{page}{1}
\pubyear{2007}
\submitted{28 Jul 2006}
\revised {10 Sep 2007}
\accepted{18 October 2007}

\maketitle

\label{firstpage}

\begin{abstract}
Building rules on top of ontologies is the ultimate goal of the logical layer of the Semantic Web. To this aim an ad-hoc mark-up language for this layer is currently under discussion. It is intended to follow the tradition of hybrid knowledge representation and reasoning systems such as $\mathcal{AL}$-log that integrates the description logic $\mathcal{ALC}$ and the function-free Horn clausal language \textsc{Datalog}. In this paper we consider the problem of automating the acquisition of these rules for the Semantic Web. We propose a general framework for rule induction that adopts the methodological apparatus of Inductive Logic Programming and relies on the expressive and deductive power of $\mathcal{AL}$-log. The framework is valid whatever the scope of induction (description vs. prediction) is. Yet, for illustrative purposes, we also discuss an instantiation of the framework which aims at description and turns out to be useful in Ontology Refinement.
\end{abstract}


\section{Introduction}

During the last decade increasing attention has been paid on \emph{ontologies} and their role in Knowledge Engineering \cite{StaabS04}. 
In the philosophical sense, we may refer to an ontology as a particular system of categories accounting for a certain vision of the world. As such, this system does not
depend on a particular language: Aristotle's ontology is always the same, independently of the language used to describe it. On the other hand, in its most prevalent use in
Artificial Intelligence, an ontology refers to an engineering artifact (more precisely, produced according to the principles of \emph{Ontological Engineering} \cite{GomezPerez04}), constituted by a specific vocabulary used to describe a certain reality, plus a set of explicit assumptions regarding the
intended meaning of the vocabulary words. This set of assumptions has usually the form of a first-order logical theory, where vocabulary words appear as unary or binary
predicate names, respectively called concepts and relations. In the simplest case, an ontology describes a hierarchy of concepts related by subsumption relationships; in
more sophisticated cases, suitable axioms are added in order to express other relationships between concepts and to constrain their intended interpretation.
The two readings of ontology described above are indeed related each other, but in order to solve the terminological impasse the word conceptualization is used to refer to the philosophical reading as appear in the following definition, based on \cite{Gruber93}: \textit{An ontology
is a formal explicit specification of a shared conceptualization for a domain of
interest}. Among the other things, this definition emphasizes the fact that an ontology
has to be specified in a language that comes with a formal semantics. Only by
using such a formal approach ontologies provide the machine interpretable meaning
of concepts and relations that is expected when using an ontology-based approach. Among the formalisms proposed by Ontological Engineering, the most currently used are \emph{Description Logics} (DLs) \cite{BaaderCMcGNPS03}. Note that DLs are decidable fragments of First Order Logic (FOL) that are incomparable with Horn Clausal Logic (HCL) as regards the expressive power \cite{Borgida96} and the semantics \cite{Rosati05}.

\begin{figure}[h]
\begin{center}
\setlength{\epsfxsize}{4.5in}
\centerline{\epsfbox{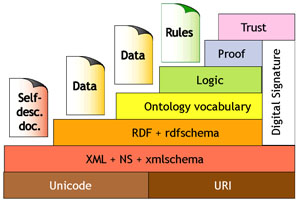}}
\caption{Architecture of the Semantic Web.}\label{fig:sw-architecture}
\end{center}
\end{figure}
Ontology Engineering, notably its DL-based approach, is playing a relevant role in the definition of the \emph{Semantic Web}. The Semantic Web is the vision of the World Wide Web enriched by machine-processable information which supports the user in his tasks \cite{Berners-Lee01}. The architecture of the Semantic Web is shown in Figure \ref{fig:sw-architecture}. It consists of several layers, each of which is equipped with an ad-hoc mark-up language.
In particular, the design of the mark-up language for the \emph{ontological layer}, OWL\footnote{\texttt{http://www.w3.org/2004/OWL/}}, has been based on the very expressive DL $\mathcal{SHOIN}(\textbf{D})$ \cite{HorrocksST00,HorrocksP-SvH03}. Whereas OWL is already undergoing the standardization process at W3C, the debate around a unified language for \emph{rules} is still ongoing. Proposals like SWRL\footnote{\texttt{http://www.w3.org/Submission/SWRL/}} extend OWL with constructs inspired to Horn clauses in order to meet the primary requirement of the \emph{logical layer}: 'to build rules on top of ontologies'. SWRL is intended to bridge the notorious gaps between DLs and HCL in a way that is similar in the spirit to hybridization in Knowledge Representation and Reasoning (KR\&R) systems such as $\mathcal{AL}$-log \cite{Donini98}. Generally speaking, \emph{hybrid systems} are KR\&R systems which are constituted by two or more subsystems dealing with distinct portions of a single knowledge base by performing specific reasoning procedures \cite{FrischC91}. The motivation for investigating and developing such systems is to improve on two basic features of KR\&R formalisms, namely \emph{representational adequacy} and \emph{deductive power}, by preserving the other crucial feature, i.e. \emph{decidability}. In particular, combining DLs with HCL can easily yield to undecidability if the interface between them is not reduced \cite{LevyR98}. The hybrid system $\mathcal{AL}$-log integrates $\mathcal{ALC}$ \cite{Schmidt-Schauss91} and \textsc{Datalog} \cite{Ceri90} by using $\mathcal{ALC}$ concept assertions essentially as type constraints on variables. It has been very recently mentioned as the blueprint for \textit{well-founded} Semantic Web rule mark-up languages because its underlying form of integration (called \textit{safe}) assures semantic and computational advantages that SWRL - though more expressive than $\mathcal{AL}$-log - currently can not assure \cite{Rosati05}. 

Defining rules (including the ones for the Semantic Web) has been usually considered as a demanding task from the viewpoint of Knowledge Engineering. It is often supported by Machine Learning algorithms that can vary in the approaches. The approach known under the name of Inductive Logic Programming (ILP) seems to be promising for the case at hand due to the common roots with Logic Programming \cite{FlachL02}. ILP has been historically concerned with rule induction from examples and background knowledge within the representation framework of HCL and with the aim of prediction \cite{Nienhuys97}. More recently ILP has moved towards either different FOL fragments (e.g., DLs) or new learning goals (e.g., description). In this paper we resort to the methodological apparatus of ILP to define a \emph{general} framework for learning rules on top of ontologies for the Semantic Web within the KR\&R framework of $\mathcal{AL}$-log. 
The framework proposed is general in the sense that it is valid whatever the scope of induction (description vs. prediction) is. For the sake of illustration we concentrate on an instantiation of the framework for the case of description.

The paper is organized as follows. Section \ref{sect:al-log} introduces the basic
notions of $\mathcal{AL}$-log. Section \ref{sect:learning-fwk} defines the framework for learning rules in $\mathcal{AL}$-log. Section
\ref{sect:inst-fwk} illustrates an instantiation of the framework. Section \ref{sect:concl} concludes the paper with final remarks. \ref{sect:owl} clarifies the links between OWL and DLs.

\section{Basics of $\mathcal{AL}$-log}\label{sect:al-log}

The system $\mathcal{AL}$-log \cite{Donini98} integrates
two KR\&R systems: Structural and relational. 

\begin{table*}[t]
\small
\caption{Syntax and semantics of $\mathcal{ALC}$.}\label{tab:ALC}
\begin{center}
\begin{tabular}{r@{\hspace{.7cm}}c@{\hspace{.7cm}}l}
\hline
\noalign{\smallskip}
bottom (resp. top) concept & $\bot$ (resp. $\top$) & $\emptyset$ (resp. $\Delta^\mathcal{I}$) \\
atomic concept & $A$ & $A^\mathcal{I} \subseteq \Delta^\mathcal{I}$\\
role & $R$ & $R^\mathcal{I} \subseteq \Delta^\mathcal{I} \times \Delta^\mathcal{I}$\\
individual & $a$ & $a^\mathcal{I} \in \Delta^\mathcal{I}$\\
\hline
\noalign{\smallskip}
concept negation & $\neg C$ & $\Delta^\mathcal{I} \setminus C^\mathcal{I}$  \\
concept conjunction & $C \sqcap D$ & $C^\mathcal{I} \cap D^\mathcal{I}$\\
concept disjunction & $C \sqcup D$ & $C^\mathcal{I} \cup D^\mathcal{I}$\\
value restriction & $\forall R.C$ & $\{x \in \Delta^\mathcal{I} \mid \forall y\ (x,y)\in R^\mathcal{I} \rightarrow y \in
C^\mathcal{I}\}$\\
existential restriction & $\exists R.C$ & $\{x \in \Delta^\mathcal{I} \mid \exists y\ (x,y)\in R^\mathcal{I} \wedge y \in
C^\mathcal{I}\}$\\
\hline
\noalign{\smallskip}
equivalence axiom & $C \equiv D$ & $C^\mathcal{I} = D^\mathcal{I}$  \\
subsumption axiom & $C \sqsubseteq D$ & $C^\mathcal{I} \subseteq D^\mathcal{I}$\\
\hline
\noalign{\smallskip}
concept assertion & $a:C$ & $a^\mathcal{I} \in C^\mathcal{I}$\\
role assertion & $\left\langle a,b \right\rangle:R$ & $(a^\mathcal{I}, b^\mathcal{I}) \in R^\mathcal{I}$\\
\hline
\end{tabular}
\end{center}
\end{table*}

\subsection{The structural subsystem}

The structural part $\Sigma$ is based on $\mathcal{ALC}$ \cite{Schmidt-Schauss91} and allows for the specification of knowledge in terms of classes (\emph{concepts}), binary relations between classes (\emph{roles}), and
instances (\emph{individuals}). Complex concepts can be defined from atomic concepts and roles by means of constructors (see Table \ref{tab:ALC}). Also $\Sigma$ can state both is-a relations between concepts (\emph{axioms}) and instance-of
relations between individuals (resp. couples of individuals) and concepts (resp. roles) (\emph{assertions}). 
An \emph{interpretation}
$\mathcal{I}=(\Delta^\mathcal{I}, \cdot^\mathcal{I})$ for $\Sigma$ consists of a domain $\Delta^\mathcal{I}$ and a mapping function
$\cdot^\mathcal{I}$. In particular, individuals are mapped to elements of
$\Delta^\mathcal{I}$ such that $a^\mathcal{I} \neq b^\mathcal{I}$ if $a \neq b$ ({\em Unique Names Assumption} (UNA) \cite{Reiter80b}). If $\mathcal{O} \subseteq \Delta^\mathcal{I}$ and $\forall a \in \mathcal{O}: a^\mathcal{I}=a$, $\mathcal{I}$ is called \emph{$\mathcal{O}$-interpretation}. 
Also $\Sigma$ represents many different interpretations, i.e. all its models (\emph{Open World Assumption} (OWA) \cite{BaaderCMcGNPS03}).

The main reasoning task for $\Sigma$ is the \emph{consistency check}. This test is performed with a \emph{tableau calculus} that starts with the tableau branch
$S = \Sigma$ and adds assertions to $S$ by means of
\emph{propagation rules} such as
\begin{itemize}
  \item $S \rightarrow_{\sqcup} S \cup \{s:D \}$ if
        \begin{enumerate}
          \item $s:C_{1} \sqcup C_{2}$ is in $S$,
          \item $D=C_{1}$ and $D=C_{2}$,
          \item neither $s:C_{1}$ nor $s:C_{2}$ is in $S$
        \end{enumerate}
  \item $S \rightarrow_{\forall} S \cup \{t:C \}$ if
        \begin{enumerate}
          \item $s:\forall R.C$ is in $S$,
          \item $sRt$ is in $S$, 
          \item $t:C$ is not in $S$
        \end{enumerate}
  \item $S \rightarrow_{\sqsubseteq} S \cup \{s:C^{\prime} \sqcup D\}$ if
        \begin{enumerate}
          \item $C\sqsubseteq D$ is in $S$,
         \item $s$ appears in $S$,
          \item $C^{\prime}$ is the NNF concept equivalent to $\neg C$
          \item $s:\neg C \sqcup D$ is not in $S$
        \end{enumerate}
  \item $S \rightarrow_{\bot} \{s:\bot \}$ if
        \begin{enumerate}
          \item $s:A$ and $s:\neg A$ are in $S$, or
          \item $s:\neg\top$ is in $S$,
          \item $s:\bot$ is not in $S$
        \end{enumerate}
\end{itemize}
until either a contradiction is generated or an interpretation
satisfying $S$ can be easily obtained from it.

\subsection{The relational subsystem}
The relational part of $\mathcal{AL}$-log allows one to define \textsc{Datalog}\footnote{For the sake of brevity we assume the reader to be familiar with \textsc{Datalog}.} programs enriched with \emph{constraints} of the form $s:C$ where $s$ is either a
constant or a variable, and $C$ is an $\mathcal{ALC}$-concept. Note that the usage of
concepts as typing constraints applies only to variables and constants that already
appear in the clause. The symbol \& separates constraints from \textsc{Datalog} atoms
in a clause.
\begin{definition}\label{def:constr-datalog-clause}
A \emph{constrained \textsc{Datalog} clause} is an implication of the form
\begin{center}
$\alpha_{0} \leftarrow \alpha_{1},\ldots, \alpha_{m} \& \gamma_{1},\ldots, \gamma_{n}$
\end{center}
where $m \geq 0$, $n \geq 0$, $\alpha_{i}$ are \textsc{Datalog} atoms and $\gamma_{j}$
are constraints. A \emph{constrained \textsc{Datalog} program} $\Pi$ is a set of
constrained \textsc{Datalog} clauses.
\end{definition}
An \emph{$\mathcal{AL}$-log knowledge base} $\mathcal{B}$ is the pair
$\langle\Sigma,\Pi\rangle$ where $\Sigma$ is an $\mathcal{ALC}$ knowledge base and
$\Pi$ is a constrained \textsc{Datalog} program. For a knowledge base to be acceptable,
it must satisfy the following conditions:
\begin{itemize}
  \item The set of \textsc{Datalog} predicate symbols appearing in $\Pi$ is disjoint
  from the set of concept and role symbols appearing in $\Sigma$.
  \item The alphabet of constants in $\Pi$ coincides with the
  alphabet $\mathcal{O}$ of the individuals in $\Sigma$.
  Furthermore, every constant in $\Pi$ appears also in $\Sigma$.
  \item For each clause in $\Pi$, each variable occurring in the constraint part occurs also in the \textsc{Datalog} part.
\end{itemize}
These properties state a \emph{safe} interaction between the structural and the relational part of an
$\mathcal{AL}$-log knowledge base, thus solving the semantic mismatch between the OWA of $\mathcal{ALC}$ and the CWA of \textsc{Datalog} \cite{Rosati05}.
This interaction is also at the basis of a model-theoretic semantics
for $\mathcal{AL}$-log. We call $\Pi_{D}$ the set of \textsc{Datalog} clauses obtained
from the clauses of $\Pi$ by deleting their constraints. We define an
\emph{interpretation} $\mathcal{J}$ for $\mathcal{B}$ as the union of an
$\mathcal{O}$-interpretation $\mathcal{I}_{\mathcal{O}}$ for $\Sigma$ (i.e. an
interpretation compliant with the unique names assumption) and an Herbrand
interpretation $\mathcal{I}_{\mathcal{H}}$ for $\Pi_{D}$. An interpretation
$\mathcal{J}$ is a \emph{model} of $\mathcal{B}$ if $\mathcal{I}_{\mathcal{O}}$ is a
model of $\Sigma$, and for each ground instance $\alpha_{0}^\prime \leftarrow \alpha_{1}^\prime,\ldots, \alpha_{m}^\prime \&
\gamma^\prime_{1},\ldots, \gamma^\prime_{n}$ of each clause $\alpha_{0} \leftarrow \alpha_{1},\ldots, \alpha_{m} \&
\gamma^\prime_{1},\ldots, \gamma^\prime_{n}$ in $\Pi$, either there exists one $\gamma^\prime_{i}$,
$i \in \{1, \ldots, n\}$, that is not satisfied by $\mathcal{J}$, or
$\alpha_{0}^\prime \leftarrow \alpha_{1}^\prime,\ldots, \alpha_{m}^\prime$ is satisfied by $\mathcal{J}$. The notion of \emph{logical
consequence} paves the way to the definition of answer set for queries. \emph{Queries}
to $\mathcal{AL}$-log knowledge bases are special cases of Definition
\ref{def:constr-datalog-clause}. An \emph{answer} to the query $Q$ is a ground
substitution $\sigma$ for the variables in $Q$. The answer $\sigma$ is \emph{correct}
w.r.t. a $\mathcal{AL}$-log knowledge base $\mathcal{B}$ if $Q\sigma$ is a logical
consequence of $\mathcal{B}$ ($\mathcal{B} \models Q\sigma$). The \emph{answer set} of
$Q$ in $\mathcal{B}$ contains all the correct answers to $Q$ w.r.t. $\mathcal{B}$.

Reasoning for $\mathcal{AL}$-log knowledge bases is based on \emph{constrained
SLD-resolution} \cite{Donini98}, i.e. an extension of SLD-resolution to deal with
constraints. In particular, the constraints of the resolvent of a query $Q$ and a
constrained \textsc{Datalog} clause $E$ are recursively simplified by replacing
couples of constraints $t:C$, $t:D$ with the equivalent constraint $t:C\sqcap D$.
The one-to-one mapping between constrained SLD-derivations and the SLD-derivations
obtained by ignoring the constraints is exploited to extend known results for
\textsc{Datalog} to $\mathcal{AL}$-log. Note that in $\mathcal{AL}$-log a derivation of
the empty clause with associated constraints does not represent a refutation. It
actually infers that the query is true in those models of $\mathcal{B}$ that satisfy
its constraints. Therefore in order to answer a query it is necessary to collect enough
derivations ending with a constrained empty clause such that every model of
$\mathcal{B}$ satisfies the constraints associated with the final query of at least one
derivation.
\begin{definition}\label{def:constrained-refutation}
Let $Q^{(0)}$ be a query $\leftarrow \beta_{1},\ldots, \beta_{m} \& \gamma_{1},\ldots,
\gamma_{n}$ to a $\mathcal{AL}$-log knowledge base $\mathcal{B}$ . A \emph{constrained
SLD-refutation} for $Q^{(0)}$ in $\mathcal{B}$ is a finite set
$\{d_{1},\ldots,d_{s}\}$ of constrained SLD-derivations for $Q^{(0)}$ in $\mathcal{B}$
such that:
\begin{enumerate}
  \item for each derivation $d_{i}$, $1 \leq i \leq s$, the last query $Q^{(n_{i})}$ of $d_{i}$ is a constrained empty clause;
  \item for every model $\mathcal{J}$ of $\mathcal{B}$, there exists at least one derivation $d_{i}$, $1 \leq i \leq s$, such that $\mathcal{J}\models Q^{(n_{i})}$
\end{enumerate}
\end{definition}
Constrained SLD-refutation is a complete and sound method for answering \emph{ground}
queries \cite{Donini98}. 
\begin{lemma}\label{lemma:ground-query-answering}
Let $Q$ be a ground query to an $\mathcal{AL}$-log knowledge base $\mathcal{B}$. It
holds that $\mathcal{B}\vdash Q$ if and only if $\mathcal{B}\models Q$.
\end{lemma}
An answer $\sigma$ to a query $Q$ is a \emph{computed answer} if there exists
a constrained SLD-refutation for $Q\sigma$ in $\mathcal{B}$ ($\mathcal{B}\vdash
Q\sigma$). The set of computed answers is called the \emph{success set} of $Q$ in
$\mathcal{B}$. Furthermore, given \emph{any} query $Q$, the success set of $Q$ in
$\mathcal{B}$ coincides with the answer set of $Q$ in $\mathcal{B}$. This provides an
operational means for computing correct answers to queries. Indeed, it is
straightforward to see that the usual reasoning methods for \textsc{Datalog} allow us
to collect in a finite number of steps enough constrained SLD-derivations for $Q$ in
$\mathcal{B}$ to construct a refutation - if any. Derivations must satisfy both
conditions of Definition \ref{def:constrained-refutation}. In particular, the latter
requires some reasoning on the structural component of $\mathcal{B}$. This is done by
applying the tableau calculus as shown in the following example.

Constrained SLD-resolution is \emph{decidable} \cite{Donini98}. Furthermore, because of the safe interaction between $\mathcal{ALC}$ and \textsc{Datalog}, it supports a form of \emph{closed world reasoning}, i.e. it allows one to pose queries under the assumption that part of the knowledge base is complete \cite{Rosati05}.

\section{The general framework for learning rules in $\mathcal{AL}$-log}\label{sect:learning-fwk}

In our framework for learning in $\mathcal{AL}$-log we represent inductive hypotheses as constrained \textsc{Datalog} clauses and data as an $\mathcal{AL}$-log knowledge base $\mathcal{B}$. In particular $\mathcal{B}$ is composed of a \emph{background knowledge} $\mathcal{K}$ and a set $O$ of \emph{observations}. We assume $\mathcal{K}\cap O =\emptyset$.

To define the framework we resort to the methodological apparatus of ILP which requires the following ingredients to be chosen:
\begin{itemize}
	\item the \emph{language $\mathcal{L}$ of hypotheses}
	\item a \emph{generality order} $\succeq$ for $\mathcal{L}$ to structure the space of hypotheses
	\item a \emph{relation} to test the \emph{coverage} of hypotheses in $\mathcal{L}$ against observations in $O$ w.r.t. $\mathcal{K}$
\end{itemize}

The framework is \textbf{general}, meaning that it is valid whatever the scope of induction (description/prediction) is. Therefore the \textsc{Datalog} literal $q(\vec{X})$\footnote{$\vec{X}$ is a tuple of variables} in the head of hypotheses represents a concept to be either discriminated from others (\textit{discriminant induction}) or characterized (\textit{characteristic induction}).

This section collects and upgrades theoretical results published in \cite{LisiM03-aiia,LisiM03-ilp,LisiE04-ilp}.

\subsection{The language of hypotheses}\label{sect:hyp-lang}

To be suitable as language of hypotheses, constrained \textsc{Datalog} clauses must satisfy the following restrictions.

First, we impose constrained \textsc{Datalog} clauses to be linked and connected (or
range-restricted) as usual in ILP \cite{Nienhuys97}. 
\begin{definition}\label{def:linkedness}
Let $H$ be a constrained \textsc{Datalog} clause. A term $t$ in some literal $l_{i}
\in H$ is \emph{linked} with linking-chain of length 0, if $t$ occurs in $head(H)$,
and is linked with linking-chain of length $d+1$, if some other term in $l_{i}$ is
linked with linking-chain of length $d$. The link-depth of a term $t$ in some $l_{i}
\in H$ is the length of the shortest linking-chain of $t$. A literal $l_{i} \in H$ is
linked if at least one of its terms is linked. The clause $H$ itself is linked if each
$l_{i} \in H$ is linked. The clause $H$ is \emph{connected} if each variable occurring
in $head(H)$ also occur in $body(H)$.
\end{definition}

Second, we impose constrained
\textsc{Datalog} clauses to be compliant with the bias of
Object Identity (OI) \cite{Semeraro98}. This bias can be considered as an extension of the UNA from the semantic level to the syntactic one of $\mathcal{AL}$-log. We would like to remind the reader that
this assumption holds in $\mathcal{ALC}$. Also it holds naturally for ground
constrained \textsc{Datalog} clauses because the semantics of $\mathcal{AL}$-log adopts
Herbrand models for the \textsc{Datalog} part and $\mathcal{O}$-models for the
constraint part. Conversely it is not guaranteed in the case of non-ground constrained
\textsc{Datalog} clauses, e.g. different variables can be unified. The OI bias can be the starting point for the definition of
either an equational theory or a quasi-order for constrained \textsc{Datalog}
clauses. The latter option relies on a restricted form of substitution whose bindings
avoid the identification of terms. 
\begin{definition}\label{def:oi-subst}
A substitution $\sigma$ is an {\em OI-substitution}
w.r.t.\ a set of terms $T$ iff $\forall t_1, t_2 \in T\!:\ t_1 \neq t_2$ yields that
$t_1\sigma \neq t_2\sigma$. 
\end{definition} 
From now on, we assume that substitutions are OI-compliant.

\subsection{The generality relation}\label{sect:gen-rel}

In ILP the key mechanism is \emph{generalization} intended as a search process through a
partially ordered space of hypotheses \cite{Mitchell82}.
The definition of a generality relation for constrained \textsc{Datalog} clauses can
disregard neither the peculiarities of $\mathcal{AL}$-log nor the methodological
apparatus of ILP. Therefore we rely on the reasoning mechanisms made available by $\mathcal{AL}$-log
knowledge bases and propose to adapt Buntine's generalized subsumption \cite{Buntine88} to our framework as follows.
\begin{definition}\label{def:coverage}
Let $H$ be a constrained \textsc{Datalog} clause, $\alpha$ a ground \textsc{Datalog} atom, and
$\mathcal{J}$ an interpretation. We say that $H$ \emph{covers} $\alpha$ under
$\mathcal{J}$ if there is a ground substitution $\theta$ for $H$ ($H\theta$ is ground)
such that $body(H)\theta$ is true under $\mathcal{J}$ and $head(H)\theta=\alpha$.
\end{definition}
\begin{definition}\label{def:B-subsumption}
Let $H_1$, $H_2$ be two constrained \textsc{Datalog} clauses and $\mathcal{B}$ an
$\mathcal{AL}$-log knowledge base. We say that $H_1$ \emph{$\mathcal{B}$-subsumes} $H_2$
if for every model $\mathcal{J}$ of $\mathcal{B}$ and every ground atom $\alpha$ such
that $H_2$ covers $\alpha$ under $\mathcal{J}$, we have that $H_1$ covers $\alpha$ under
$\mathcal{J}$.
\end{definition}

We can define a generality relation $\succeq_{\mathcal{B}}$ for constrained
\textsc{Datalog} clauses on the basis of $\mathcal{B}$-subsumption. It can be easily
proven that $\succeq_{\mathcal{B}}$ is a quasi-order (i.e. it is a reflexive and
transitive relation) for constrained \textsc{Datalog} clauses.
\begin{definition}\label{def:generality-under-B-subsumption}
Let $H_1$, $H_2$ be two constrained \textsc{Datalog} clauses and $\mathcal{B}$ an
$\mathcal{AL}$-log knowledge base. We say that $H_1$ is \emph{at least as general as}
$H_2$ under $\mathcal{B}$-subsumption, $H_1 \succeq_{\mathcal{B}} H_2$, iff $H_1$
$\mathcal{B}$-subsumes $H_2$. Furthermore, $H_1$ is \emph{more general than} $H_2$ under
$\mathcal{B}$-subsumption, $H_1 \succ_{\mathcal{B}} H_2$, iff $H_1 \succeq_{\mathcal{B}} H_2$
and $H_{2} \not\succeq_{\mathcal{B}} H_1$. Finally, $H_1$ is \emph{equivalent} to $H_2$ under
$\mathcal{B}$-subsumption, $H_1 \sim_{\mathcal{B}} H_2$, iff $H_1 \succeq_{\mathcal{B}} H_2$
and $H_2 \succeq_{\mathcal{B}} H_1$.
\end{definition}

The next lemma shows the definition of $\mathcal{B}$-subsumption to be equivalent
to another formulation, which will be more convenient in later proofs than the
definition based on covering.
\begin{definition}\label{def:Skolem-subst}
Let $\mathcal{B}$ be an $\mathcal{AL}$-log knowledge base and $H$ be a constrained
\textsc{Datalog} clause. Let $X_1,\ldots,X_n$ be all the variables appearing in $H$,
and $a_1,\ldots,a_n$ be distinct constants (individuals) not appearing in
$\mathcal{B}$ or $H$. Then the substitution $\{X_1/a_1,\ldots,X_n/a_n\}$ is called a
\emph{Skolem substitution} for $H$ w.r.t. $\mathcal{B}$.
\end{definition}
\begin{lemma}\label{lemma:B-subsumption1}
Let $H_1$, $H_2$ be two constrained \textsc{Datalog} clauses, $\mathcal{B}$ an
$\mathcal{AL}$-log knowledge base, and $\sigma$ a Skolem substitution for $H_2$ with
respect to $\{H_1\} \cup \mathcal{B}$. We say that $H_1 \succeq_{\mathcal{B}} H_2$ iff there
exists a ground substitution $\theta$ for $H_1$ such that (i)
$head(H_1)\theta=head(H_2)\sigma$ and (ii) $\mathcal{B} \cup body(H_2)\sigma \models
body(H_1)\theta$.
\begin{proof}
\begin{description}
  \item[$(\Rightarrow)$] Suppose $H_1 \succeq_{\mathcal{B}} H_2$. Let $\mathcal{B}^\prime$
be the knowledge base $\mathcal{B} \cup body(H_2)\sigma$ and $\mathcal{J}=\langle
\mathcal{I}_\mathcal{O}, \mathcal{I}_H\rangle$ be a model of $\mathcal{B}^\prime$
where $\mathcal{I}_\mathcal{O}$ is the minimal $\mathcal{O}$-model of $\Sigma$ and
$\mathcal{I}_H$ be the least Herbrand model of the \textsc{Datalog} part of
$\mathcal{B}^\prime$. The substitution $\sigma$ is a ground substitution for $H_2$, and
$body(H_2)\sigma$ is true under $\mathcal{J}$, so $H_2$ covers $head(H_2)\sigma$ under
$\mathcal{J}$ by Definition \ref{def:coverage}. Then $H_1$ must also cover
$head(H_2)\sigma$ under $\mathcal{J}$. Thus there is a ground substitution $\theta$ for
$H_1$ such that $head(H_1)\theta=head(H_2)\sigma$, and $body(H_1)\theta$ is true under
$\mathcal{J}$, i.e. $\mathcal{J} \models body(H_1)\theta$. By properties of the least
Herbrand model, it holds that $\mathcal{B} \cup body(H_2)\sigma \models \mathcal{J}$,
hence $\mathcal{B} \cup body(H_2)\sigma \models body(H_1)\theta$.
  \item[$(\Leftarrow)$] Suppose there is a ground substitution $\theta$ for $H_1$, such
  that $head(H_1)\theta=head(H_2)\sigma$ and $\mathcal{B} \cup body(H_2)\sigma \models
body(H_1)\theta$. Let $\alpha$ be some ground atom and $\mathcal{J}_\alpha$ some model
of $\mathcal{B}$ such that $H_2$ covers $\alpha$ under $\mathcal{J}_\alpha$. To prove
that $H_1 \succeq_{\mathcal{B}} H_2$ we need to prove that $H_1$ covers $\alpha$ under
$\mathcal{J}_\alpha$.

Construct a substitution $\theta^\prime$ from $\theta$ as follows: for every binding
$X /c \in \sigma$, replace $c$ in bindings in $\theta$ by $X$. Then we have
$H_1 \theta^\prime \sigma= H_1 \theta$ and none of the Skolem constants of $\sigma$ occurs in
$\theta^\prime$. Then $head(H_1)\theta^\prime \sigma = head(H_1)\theta = head(H_2)\sigma$,
so $head(H_1)\theta^\prime = head(H_2)$. Since $H_2$ covers $\alpha$ under
$\mathcal{J}_\alpha$, there is a ground substitution $\gamma$ for $H_2$, such that
$body(H_2)\gamma$ is true under $\mathcal{J}_\alpha$, and $head(H_2)\gamma=\alpha$. This
implies that $head(H_1)\theta^\prime \gamma = head(H_2)\gamma=\alpha$.

It remains to show that $body(H_1)\theta^\prime \gamma$ is true under
$\mathcal{J}_\alpha$. Since $\mathcal{B} \cup body(H_2)\sigma \models
body(H_1)\theta^\prime \sigma$ and $\leftarrow body(H_1)\theta^\prime \sigma$ is a ground
query, it follows from Lemma \ref{lemma:ground-query-answering} that there exists a
constrained SLD-refutation for $\leftarrow body(H_1)\theta^\prime \sigma$ in
$\mathcal{B} \cup body(H_2)\sigma$. By Definition \ref{def:constrained-refutation} there
exists a finite set $\{d_{1},\ldots,d_{m}\}$ of constrained SLD-derivations, having
$\leftarrow body(H_1)\theta^\prime \sigma$ as top clause and elements of $\mathcal{B}
\cup body(H_2)\sigma$ as input clauses, such that for each derivation $d_{i}$, $i \in
\{1,\ldots,m\}$, the last query $Q^{(n_{i})}$ of $d_{i}$ is a constrained empty clause
and for every model $\mathcal{J}$ of $\mathcal{B} \cup body(H_2)\sigma$, there exists at
least one derivation $d_{i}$, $i \in \{1,\ldots,m\}$, such that $\mathcal{J}\models
Q^{(n_{i})}$. We want to turn this constrained SLD-refutation for $\leftarrow
body(H_1)\theta^\prime \sigma$ in $\mathcal{B} \cup body(H_2)\sigma$ into a constrained
SLD-refutation for $\leftarrow body(H_1)\theta^\prime \gamma$ in $\mathcal{B} \cup
body(H_2)\gamma$, thus proving that $\mathcal{B} \cup body(H_2)\gamma \models
body(H_1)\theta^\prime \gamma$. Let $X_1, \ldots, X_n$ be the variables in $body(H_2)$ such that $\{X_1 /c_1, \ldots, X_n /c_n \}\subseteq \sigma$, and $\{X_1 /t_1, \ldots, X_n /t_n
\}\subseteq \gamma$. If we replace each Skolem constant $c_j$ by $t_j$, $1\leq j\leq
n$, in both the SLD-derivations and the models of $\mathcal{B} \cup body(H_2)\sigma$ we
obtain a constrained SLD-refutation of $body(H_1)\theta^\prime \gamma$ in $\mathcal{B}
\cup body(H_2)\gamma$. Hence $\mathcal{B} \cup body(H_2)\gamma \models
body(H_1)\theta^\prime \gamma$. Since $\mathcal{J}_\alpha$ is a model of $\mathcal{B}
\cup body(H_2)\gamma$, it is also a model of $body(H_1)\theta^\prime \gamma$.
\end{description}
\end{proof}
\end{lemma}

The relation between $\mathcal{B}$-subsumption and constrained SLD-resolution is given
below. It provides an operational means for checking $\mathcal{B}$-subsumption.
\begin{thm}\label{thm:B-subsumption-test}
Let $H_1$, $H_2$ be two constrained \textsc{Datalog} clauses, $\mathcal{B}$ an
$\mathcal{AL}$-log knowledge base, and $\sigma$ a Skolem substitution for $H_2$ with
respect to $\{H_1\} \cup \mathcal{B}$. We say that $H_1 \succeq_{\mathcal{B}} H_2$ iff there
exists a substitution $\theta$ for $H_1$ such that (i) $head(H_1)\theta=head(H_2)$ and (ii)
$\mathcal{B} \cup body(H_2)\sigma \vdash body(H_1)\theta\sigma$ where
$body(H_1)\theta\sigma$ is ground.
\begin{proof}
By Lemma \ref{lemma:B-subsumption1}, we have $H_1 \succeq_{\mathcal{B}} H_2$ iff there
exists a ground substitution $\theta^\prime$ for $H_1$, such that
$head(H_1)\theta^\prime=head(H_2)\sigma$ and $\mathcal{B} \cup body(H_2)\sigma \models
body(H_1)\theta^\prime$. Since $\sigma$ is a Skolem substitution, we can define a
substitution $\theta$ such that $H_1 \theta\sigma = H_1 \theta^\prime$ and none of the Skolem
constants of $\sigma$ occurs in $\theta$. Then $head(H_1)\theta=head(H_2)$ and
$\mathcal{B} \cup body(H_2)\sigma \models body(H_1)\theta\sigma$. Since
$body(H_1)\theta\sigma$ is ground, by Lemma \ref{lemma:ground-query-answering} we have
$\mathcal{B} \cup body(H_2)\sigma \vdash body(H_1)\theta\sigma$, so the thesis follows.
\end{proof}
\end{thm}

The decidability of $\mathcal{B}$-subsumption follows from the decidability of both
generalized subsumption in \textsc{Datalog} \cite{Buntine88} and query answering in
$\mathcal{AL}$-log \cite{Donini98}.

\subsection{Coverage relations}\label{sect:cover-test}

When defining coverage relations we make assumptions as regards the representation of observations because it impacts the definition of coverage. In ILP there are two choices: we can represent an observation as either a ground definite clause or a set of ground unit clauses. The former is peculiar to the normal ILP setting (also called \emph{learning from
implications}) \cite{FrazierP93}, whereas the latter is usual in the logical setting of \emph{learning from interpretations} \cite{DeRaedtD94}. The representation choice for observations and the scope of induction are orthogonal dimensions as clearly explained in \cite{DeRaedt97}. Therefore we prefer the term 'observation' to the term 'example' for the sake of generality.

In the logical setting of \emph{learning from entailment} extended to $\mathcal{AL}$-log, an observation $o_i \in O$ is represented as a ground constrained \textsc{Datalog} clause having a ground atom $q(\vec{a}_i)$\footnote{$\vec{a}_i$ is a tuple of constants} in the head. 

\begin{definition}\label{def:coverage_ent}
Let $H \in \mathcal{L}$ be a hypothesis, $\mathcal{K}$ a background knowledge and $o_i \in O$ an observation. We say that \emph{$H$ covers $o_i$ under entailment} w.r.t $\mathcal{K}$ iff $\mathcal{K} \cup H \models o_i$. 
\end{definition}

In order to provide an operational means for testing this coverage relation we resort to the Deduction Theorem for first-order logic formulas \cite{Nienhuys97}.
\begin{thm}\label{thm:ded-thm}
Let $\Sigma$ be a set of formulas, and $\phi$ and $\psi$ be formulas. We say that $\Sigma \cup \{\phi\} \models \psi$ iff $\Sigma \models (\phi \rightarrow \psi)$.
\end{thm}
\begin{thm}\label{thm:coverage-test1}
Let $H\in \mathcal{L}$ be a hypothesis, $\mathcal{K}$ a background knowledge, and $o_i \in O$ an observation. We say that \emph{$H$ covers $o_i$ under entailment w.r.t. $\mathcal{K}$} iff $\mathcal{K} \cup body(o_i) \cup H \vdash q(\vec{a}_i)$.
\begin{proof}
The following chain of equivalences holds:
\begin{itemize}
	\item $H$ covers $o_i$ under entailment w.r.t. $\mathcal{K}$ $\leftrightarrow$ (by Definition \ref{def:coverage_ent})
	\item $\mathcal{K} \cup H \models q(\vec{a}_i) \leftarrow body(o_i)$ $\leftrightarrow$ (by Theorem \ref{thm:ded-thm})
	\item $\mathcal{K} \cup H \cup body(o_i) \models q(\vec{a}_i)$ $\leftrightarrow$ (by	Lemma \ref{lemma:ground-query-answering})
	\item $\mathcal{K} \cup body(o_i) \cup H \vdash q(\vec{a}_i)$  
\end{itemize}
\end{proof}
\end{thm}

In the logical setting of \emph{learning from interpretations} extended to $\mathcal{AL}$-log, an observation $o_i \in O$ is represented as a couple $(q(\vec{a}_i), \mathcal{A}_i)$ where $\mathcal{A}_i$ is a set containing ground \textsc{Datalog} facts concerning the individual $i$. 

\begin{definition}\label{def:coverage_int}
Let $H \in \mathcal{L}$ be a hypothesis, $\mathcal{K}$ a background knowledge and $o_i \in O$ an observation. We say that \emph{$H$ covers $o_i$ under interpretations w.r.t. $\mathcal{K}$} iff $\mathcal{K} \cup \mathcal{A}_i \cup H \models q(\vec{a}_i)$. 
\end{definition} 

\begin{thm}\label{thm:coverage-test2}
Let $H \in \mathcal{L}$ be a hypothesis, $\mathcal{K}$ a background knowledge, and $o_i \in O$ an observation. We say that \emph{$H$ covers $o_i$ under interpretations w.r.t. $\mathcal{K}$} iff $\mathcal{K} \cup \mathcal{A}_i \cup H \vdash q(\vec{a}_i)$.
\begin{proof}
Since $q(\vec{a}_i)$ is a ground query to the $\mathcal{AL}$-log knowledge base $\mathcal{B}=\mathcal{K} \cup \mathcal{A}_i \cup H$, the thesis follows from Definition \ref{def:coverage_int} and Lemma \ref{lemma:ground-query-answering}.
\end{proof}
\end{thm}

Note that both coverage tests can be reduced to query answering.


\section{Instantiating the framework for Ontology Refinement}\label{sect:inst-fwk}

\emph{Ontology Refinement} is a phase in the Ontology Learning process that aims at the adaptation of an existing ontology to a specific domain or the needs of a particular user \cite{MaedcheS04}. 
In this section we consider the problem of Concept Refinement which is about refining a known concept, called \emph{reference concept}, belonging to an existing taxonomic ontology in the light of new knowledge coming from a relational data source. A taxonomic ontology is an ontology organized around the is-a relationship between concepts \cite{GomezPerez04}. We assume that a \emph{concept} $\mathcal{C}$ consists of two parts: an \emph{intension} $int(\mathcal{C})$ and an \emph{extension} $ext(\mathcal{C})$. The former is an expression belonging to a logical language $\mathcal{L}$ whereas the latter is a set of objects that satisfy the former. 
More formally, given
\begin{itemize}
  \item a taxonomic ontology $\Sigma$, 
  \item a reference concept $C_{ref} \in \Sigma$, 
  \item a relational data source $\Pi$,
  \item a logical language $\mathcal{L}$
\end{itemize}
the \emph{Concept Refinement} problem is to find a taxonomy $\mathcal{G}$ of concepts $\mathcal{C}_i$ such that (i) $int(\mathcal{C}_i) \in \mathcal{L}$ and (ii) $ext(\mathcal{C}_i) \subset ext(C_{ref})$. Therefore $\mathcal{G}$ is structured according to the \emph{subset relation} between concept extensions. Note that $C_{ref}$ is among both the concepts defined in $\Sigma$ and the symbols of $\mathcal{L}$. Furthermore $ext(\mathcal{C}_i)$ relies on notion of satisfiability of $int(\mathcal{C}_i)$ w.r.t. $\mathcal{B}=\Sigma \cup \Pi$. We would like to emphasize that $\mathcal{B}$ includes $\Sigma$ because in Ontology Refinement, as opposite to other forms of Ontology Learning such as Ontology Extraction (or Building), it is mandatory to consider the existing ontology and its existing connections. Thus, a formalism like $\mathcal{AL}$-log suits very well the hybrid nature of $\mathcal{B}$ (see Section \ref{sect:kr}).
 
In our ILP approach the Ontology Refinement problem at hand is reformulated as a Concept Formation problem \cite{LisiE07-ilp}. \emph{Concept Formation} indicates a ML task that refers to the acquisition of conceptual hierarchies in which each concept has a flexible, non-logical definition and in which learning occurs incrementally and without supervision \cite{Langley87}. More precisely, it is to take a large number of unlabeled training instances: to find clusterings that group those instances in categories: to find an intensional definition for each category that summarized its instances; and to find a hierarchical organization for those categories \cite{GennariLF89}. Concept Formation stems from \emph{Conceptual Clustering} \cite{MichalskiS83}. The two differ substantially in the methods: The latter usually applies bottom-up batch algorithms whereas the former prefers top-down incremental ones. Yet the methods are similar in the scope of induction, i.e. \emph{prediction}, as opposite to (Statistical) Clustering \cite{Hartigan01} and Frequent Pattern Discovery \cite{Mannila97} whose goal is to describe a data set.
According to the commonly accepted formulation of the task \cite{Langley87,GennariLF89}, Concept Formation can be decomposed in two sub-tasks:
\begin{enumerate}
	\item clustering
	\item characterization
\end{enumerate}
The former consists of using internalised heuristics to organize the observations into categories whereas the latter consists in determining a concept (that is, an intensional description) for each extensionally defined subset discovered by clustering.
We propose a pattern-based approach for the former (see Section \ref{sect:phase1}) and a bias-based approach for the latter (see Section \ref{sect:phase2}). In particular, the clustering approach is pattern-based because it relies on the aforementioned commonalities between Clustering and Frequent Pattern Discovery. Descriptive tasks fit the ILP setting of \emph{characteristic induction} \cite{DeRaedtD97}. A distinguishing feature of this form of induction is the density of solution space. The setting of \emph{learning from interpretations} has been shown to be a promising way of dealing with such spaces \cite{Blockeel99}.
\begin{definition}\label{def:char-ind-from-interpr}
Let $\mathcal{L}$ be a hypothesis language, $\mathcal{K}$ a background knowledge, $O$ a set of observations, and $M(\mathcal{B})$ a model constructed from $\mathcal{B}=\mathcal{K}\cup O$. The goal of \emph{characteristic induction from interpretations} is to find a set $\mathcal{H} \subseteq \mathcal{L}$ of hypotheses such that (i) $\mathcal{H}$ is true in $M(\mathcal{B})$, and (ii) for each $H \in \mathcal{L}$, if $H$ is true in $M(\mathcal{B})$ then $\mathcal{H} \models H$.
\end{definition}
In the following subsection we will clarify the nature of $\mathcal{K}$ and $O$.

\subsection{Representation Choice}\label{sect:kr}

The KR\&R framework for conceptual knowledge in the Concept Refinement problem at hand is the one offered by $\mathcal{AL}$-log. 

The \textbf{taxonomic ontology $\Sigma$} is a $\mathcal{ALC}$ knowledge base. From now on we will call \emph{input concepts} all the concepts occurring in $\Sigma$.

\begin{figure}[ht]
\begin{center}
\setlength{\epsfxsize}{4.8in}
\centerline{\epsfbox{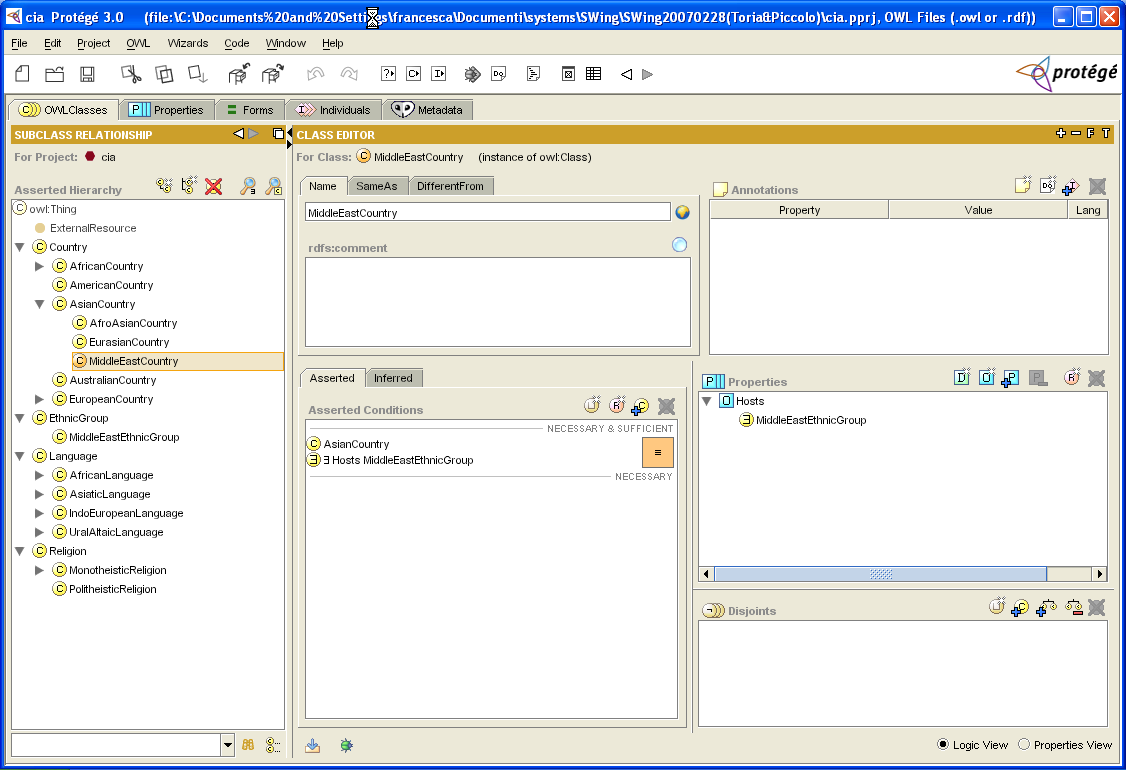}}
\caption{The ontology $\Sigma_\texttt{CIA}$ used as an example throughout Section \ref{sect:kr}.}\label{fig:cia-owl}
\end{center}
\end{figure}
\begin{example}\label{ex:cia-onto}
Throughout this Section, we will refer to the $\mathcal{ALC}$ ontology $\Sigma_\texttt{CIA}$ (see Figure \ref{fig:cia-owl}) concerning countries, ethnic groups, languages, and religions of the world, and built according to Wikipedia\footnote{\texttt{http://www.wikipedia.org/}} taxonomies. For instance, the expression
\begin{tabbing}
  MMMM \= MMMMMM \= MM \= \kill
  $\texttt{MiddleEastCountry}\equiv\texttt{AsianCountry}\sqcap\exists\texttt{Hosts.MiddleEasternEthnicGroup}$.
\end{tabbing}
is an equivance axiom that defines the concept $\texttt{MiddleEastCountry}$ as an Asian country which hosts at least one Middle Eastern ethnic group. In particular, Armenia (\texttt{'ARM'}) and Iran (\texttt{'IR'}) are two of the 15 countries that are classified as Middle Eastern.
\end{example}

The \textbf{relational data source $\Pi$} is a \textsc{Datalog} program. The extensional part of $\Pi$ is partitioned into portions $\mathcal{A}_{i}$ each of which refers to an individual $a_i$ of $C_{ref}$. The link between $\mathcal{A}_{i}$ and $a_i$ is represented with the \textsc{Datalog} literal $q(a_i)$. The pair $(q(a_i), \mathcal{A}_{i})$ is called \emph{observation}. The intensional part (IDB) of $\Pi$ together with the whole $\Sigma$ is considered as \emph{background knowledge} for the problem at hand. 

\begin{example}\label{ex:cia-db}
An $\mathcal{AL}$-log knowledge base $\mathcal{B}_\texttt{CIA}$ has been obtained by integrating $\Sigma_\texttt{CIA}$ and a \textsc{Datalog} program $\Pi_\texttt{CIA}$ based on the on-line 1996 CIA World Fact Book\footnote{\texttt{http://www.odci.gov/cia/publications/factbook/}}.
The extensional part of $\Pi_\texttt{CIA}$ consists of
\textsc{Datalog} facts\footnote{\texttt{http://www.dbis.informatik.uni-goettingen.de/
Mondial/mondial-rel-facts.flp}} grouped according to the individuals of \texttt{MiddleEastCountry}. In particular, the observation $(q(\texttt{'IR'}), \mathcal{A}_{\texttt{IR}})$ contains \textsc{Datalog} facts such as 
\begin{tabbing}
  MMMMM \= MMMMM \= MM \= \kill
\texttt{language('IR','Persian',58).}\\
\texttt{religion('IR','ShiaMuslim',89).}\\
\texttt{religion('IR','SunniMuslim',10).}
\end{tabbing}
concerning the individual \texttt{'IR'}. 
The intensional part of $\Pi_\texttt{CIA}$ defines two views on \texttt{language} and \texttt{religion}:
\begin{tabbing}
  MMMMMMM \= MMMMMMM \= MM \= \kill
  \texttt{speaks(CountryID, LanguageN)}$\leftarrow$\texttt{language(CountryID,LanguageN,Perc)}\\
  \> \> \& \texttt{CountryID:Country, LanguageN:Language.}\\
  \texttt{believes(CountryID, ReligionN)}$\leftarrow$\texttt{religion(CountryID,ReligionN,Perc)}\\
  \> \> \& \texttt{CountryID:Country, ReligionN:Religion.}
\end{tabbing}
that can deduce new \textsc{Datalog} facts when triggered on $\mathcal{B}_\texttt{CIA}$. It forms the background knowledge $\mathcal{K}_\texttt{CIA}$ together with the whole $\Sigma_\texttt{CIA}$.
\end{example}

The \textbf{language $\mathcal{L}$} contains expressions, called $\mathcal{O}$-queries, relating individuals of $C_{ref}$ to individuals of other input concepts (\emph{task-relevant concepts}). An \emph{$\mathcal{O}$-query} is a constrained \textsc{Datalog}
clause of the form 
\begin{center}$Q=q(X) \leftarrow \alpha_{1},\ldots, \alpha_{m} \&
X:C_{ref}, \gamma_{2}, \ldots, \gamma_{n}$,
\end{center} 
which is compliant with the properties mentioned in Section \ref{sect:hyp-lang}.
The $\mathcal{O}$-query
\begin{center}
$Q_t = q(X) \leftarrow \& X:C_{ref}$
\end{center}
is called \emph{trivial} for $\mathcal{L}$ because it only contains the constraint for the \emph{distinguished variable} $X$.
Furthermore, the language $\mathcal{L}$ is \emph{multi-grained}, i.e. it contains expressions at multiple levels of description granularity. Indeed it is implicitly defined by a \emph{declarative bias specification} which consists of a finite alphabet $\mathcal{A}$ of \textsc{Datalog} predicate names appearing in $\Pi$ and finite alphabets $\Gamma^l$ (one for each level $l$ of description granularity) of $\mathcal{ALC}$ concept names occurring in $\Sigma$. Note that $\alpha_{i}$'s are taken from $\mathcal{A}$ and $\gamma_{j}$'s are taken from $\Gamma^l$. We impose $\mathcal{L}$ to be finite by specifying some bounds, mainly $maxD$ for the maximum depth of search and $maxG$ for the maximum level of granularity.
\begin{example}\label{ex:sw-pat}
We want to refine the concept $\texttt{MiddleEastCountry}$ belonging to $\Sigma_\texttt{CIA}$ in the light of the new knowledge coming from $\Pi_\texttt{CIA}$. More precisely, we want to describe Middle East countries (individuals of the reference concept) with respect to the religions believed and the languages spoken (individuals of the task-relevant concepts) at three levels of granularity ($maxG=3$). To this aim we define $\mathcal{L}_\texttt{CIA}$ as the set of $\mathcal{O}$-queries with $C_{ref}= \texttt{MiddleEastCountry}$ that can be generated from the alphabet
\begin{tabbing}
  MMMM \= MM \= MM \= \kill
  $\mathcal{A}$= \{\texttt{believes/2}, \texttt{speaks/2}\}
\end{tabbing}
of \textsc{Datalog} binary predicate names, and the alphabets
\begin{tabbing}
  MMMM \= MM \= MM \= \kill
  $\Gamma^1$= \{\texttt{Language}, \texttt{Religion}\}\\
  $\Gamma^2$= $\{\texttt{IndoEuropeanLanguage}, \ldots, \texttt{MonotheisticReligion}, \ldots\}$\\
  $\Gamma^3$= $\{\texttt{IndoIranianLanguage}, \ldots, \texttt{MuslimReligion}, \ldots\}$
\end{tabbing}
of $\mathcal{ALC}$ concept names for $1 \leq l \leq 3$, up to $maxD=5$. Note that the names in $\mathcal{A}$ are taken from $\Pi_\texttt{CIA}$ whereas the names in $\Gamma^l$'s are taken from $\Sigma_\texttt{CIA}$. Examples of $\mathcal{O}$-queries in $\mathcal{L}_\texttt{CIA}$ are:
\begin{tabbing}
  MMMMMM \= MM \= MM \= \kill
  $Q_{t}$= \texttt{q(X)} $\leftarrow$ \& \texttt{X:MiddleEastCountry}\\
  $Q_{1}$= \texttt{q(X)} $\leftarrow$ \texttt{speaks(X,Y)} \& \texttt{X:MiddleEastCountry}, \texttt{Y:Language}\\
  $Q_{2}$= \texttt{q(X)} $\leftarrow$ \texttt{speaks(X,Y)} \& \texttt{X:MiddleEastCountry}, \texttt{Y:IndoEuropeanLanguage}\\
  $Q_{3}$= \texttt{q(X)} $\leftarrow$ \texttt{believes(X,Y)}\& \texttt{X:MiddleEastCountry}, \texttt{Y:MuslimReligion}
\end{tabbing}
where $Q_{t}$ is the trivial $\mathcal{O}$-query for $\mathcal{L}_\texttt{CIA}$, $Q_{1}\in \mathcal{L}_\texttt{CIA}^1$, $Q_{2} \in \mathcal{L}_\texttt{CIA}^2$, and $Q_{3} \in \mathcal{L}_\texttt{CIA}^3$.
\end{example}

\emph{Output concepts} are the concepts automatically formed out of the input ones by taking into account the relational data source. 
Thus, an output concept $\mathcal{C}$ has an $\mathcal{O}$-query $Q\in \mathcal{L}$ as intension and the set $\answerset{Q,\mathcal{B}}$ of correct answers to $Q$ w.r.t. $\mathcal{B}$ as extension. Note that this set contains the substitutions $\theta_i$'s for the distinguished variable of $Q$ such that there exists a correct answer to $body(Q)\theta_i$ w.r.t. $\mathcal{B}$. In other words, the extension is the set of individuals of $C_{ref}$ satisfying the intension. Also with reference to Section \ref{sect:cover-test} note that proving that an $\mathcal{O}$-query $Q$ covers an observation $(q(a_i), \mathcal{A}_{i})$ w.r.t. $\mathcal{K}$ equals to proving that $\theta_i = \{X / a_i \}$ is a correct answer to $Q$ w.r.t. $\mathcal{B}_{i} = \mathcal{K} \cup \mathcal{A}_{i}$. 
\begin{example}\label{ex:cia-answerset}
The output concept having $Q_1$ as intension has extension $\answerset{Q_1,\mathcal{B}_\texttt{CIA}}$ = \{\texttt{'ARM'}, \texttt{'IR'}, \texttt{'SA'}, \texttt{'UAE'}\}. In particular, 
$Q_1$ covers the observation $(\texttt{q('IR')}, \mathcal{A}_{\texttt{IR}})$ w.r.t. $\mathcal{K}_{\texttt{CIA}}$. This coverage test is equivalent to answering the query $\leftarrow$ \texttt{q('IR')} w.r.t. $\mathcal{K}_{\texttt{CIA}} \cup \mathcal{A}_{\texttt{IR}} \cup Q_1$. 
\end{example}

Output concepts are organized into a taxonomy $\mathcal{G}$ rooted in $C_{ref}$ and structured as a Directed Acyclic Graph (DAG) according to the \emph{subset relation} between concept extensions. Note that one such ordering is in line with the set-theoretic semantics of the subsumption relation in ontology languages (see, e.g., the semantics of $\sqsubseteq$ in $\mathcal{ALC}$).

\subsection{Pattern-based clustering }\label{sect:phase1}

\emph{Frequent Pattern Discovery} is about the discovery of regularities in a data set \cite{Mannila97}. A frequent pattern is an
intensional description, expressed in a language $\mathcal{L}$, of a subset of a given data set \textbf{r} whose cardinality exceeds a user-defined threshold (\emph{minimum support}). Note that patterns can refer to multiple levels of description granularity (\emph{multi-grained patterns}) \cite{HanF99}. Here \textbf{r} typically encompasses a taxonomy $\mathcal{T}$. More precisely, the problem of \emph{frequent pattern discovery at $l$ levels of description
granularity}, $1\leq l\leq maxG$, is to find the set $\mathcal{F}$ of all the frequent patterns
expressible in a multi-grained language $\mathcal{L}=\{\mathcal{L}^l\}_{1\leq l\leq maxG}$ and evaluated against \textbf{r} w.r.t. a set $\{minsup^l\}_{1\leq l\leq maxG}$ of minimum support thresholds by means of the evaluation function $supp$. In this case, $P \in \mathcal{L}^l$ with support $s$ is frequent in \textbf{r} if (i) $s\geq minsup^l$ and (ii) all ancestors of $P$ w.r.t. $\mathcal{T}$ are frequent in \textbf{r}.
The blueprint of most algorithms for frequent pattern discovery is the \emph{levelwise search} method \cite{Mannila97} which searches the space $(\mathcal{L}, \succeq)$ of patterns organized according to a generality order $\succeq$ in a breadth-first manner, starting from the most general pattern in $\mathcal{L}$ and alternating candidate generation and candidate evaluation phases. The underlying assumption is that $\succeq$ is a quasi-order monotonic w.r.t. \emph{supp}. Note that the method proposed in \cite{Mannila97} is also at the basis of algorithms for the variant of the task defined in \cite{HanF99}. 

A frequent pattern highlights a regularity in \textbf{r}, therefore it can be considered as the clue of a data cluster. Note that clusters are concepts partially specified (called \emph{emerging concepts}): only the extension is known. We propose to detect emerging concepts by applying the method of \cite{LisiM04} for frequent pattern discovery at $l$, $1\leq l\leq maxG$, levels of description granularity and $k$, $1\leq k\leq maxD$, levels of search depth. It adapts \cite{Mannila97,HanF99} to the KR\&R framework of $\mathcal{AL}$-log as follows. For $\mathcal{L}$ being a multi-grained language of $\mathcal{O}$-queries, we need to define first \emph{supp}, then $\succeq$. The \emph{support} of an $\mathcal{O}$-query $Q \in \mathcal{L}$ w.r.t. an $\mathcal{AL}$-log knowledge base $\mathcal{B}$ is defined as
\begin{center}
$supp(Q,\mathcal{B}) = \mid \answerset{Q,\mathcal{B}}
\mid / \mid \answerset{Q_t,\mathcal{B}} \mid$
\end{center}
and supplies the percentage of individuals of $C_{ref}$ that satisfy $Q$.  
\begin{example}\label{ex:cia-support}
Since $\mid\answerset{Q_t,\mathcal{B}_\texttt{CIA}}\mid=\mid\texttt{MiddleEastCountry}\mid=15$, then $supp(Q_1,\mathcal{B}_\texttt{CIA}) = 4/15 = 26.6 \%$.
\end{example}

Being a special case of constrained \textsc{Datalog} clauses, $\mathcal{O}$-queries can be ordered according to the $\mathcal{B}$-subsumption relation introduced in Section \ref{sect:gen-rel}. It has been proved that $\succeq_{\mathcal{B}}$ is a quasi-order that fulfills the condition of monotonicity w.r.t. \emph{supp} \cite{LisiM04}. Also note that the underlying reasoning mechanism of $\mathcal{AL}$-log makes
$\mathcal{B}$-subsumption more powerful than generalized subsumption as illustrated in the following example.
\begin{example}\label{ex:gen-rel3}
It can be checked that $Q_1 \succeq_\mathcal{B} Q_{2}$ by choosing $\sigma$=\{\texttt{X/a}, \texttt{Y/b}\}
as a Skolem substitution for $Q_{2}$ w.r.t. $\mathcal{B}_\texttt{CIA}\cup \{Q_1\}$ and
$\theta=\emptyset$ as a substitution for $Q_1$. Similarly it can be proved that $Q_2 \not\succeq_\mathcal{B} Q_{1}$.
Furthermore, it can be easily verified that $Q_{3}$ $\mathcal{B}$-subsumes the following $\mathcal{O}$-query in $\mathcal{L}^{3}_\texttt{CIA}$
\begin{tabbing}
  MMMMMM \= MM \= MM \= \kill
  $Q_{4}$= \texttt{q(A)} $\leftarrow$ \texttt{believes(A,B)}, \texttt{believes(A,C)}\&\\
  \> \texttt{A:MiddleEastCountry}, \texttt{B:MuslimReligion}
\end{tabbing}
by choosing $\sigma$=\{\texttt{A/a}, \texttt{B/b},
\texttt{C/c}\} as a Skolem substitution for $Q_{4}$ w.r.t. $\mathcal{B}_\texttt{CIA} \cup
\{Q_3\}$ and $\theta$=\{\texttt{X/A}, \texttt{Y/B}\} as a substitution for $Q_3$. 
Note that $Q_4 \not\succeq_\mathcal{B} Q_3$ under the
OI bias. Indeed this bias does not admit the substitution
\{\texttt{A/X}, \texttt{B/Y}, \texttt{C/Y}\} for $Q_4$ which would make
it possible to verify conditions (i) and (ii) of the $\succeq_{\mathcal{B}}$ test.
\end{example}

We would like to emphasize that $\Sigma$, besides contributing to the definition of $\mathcal{L}$ (see Section \ref{sect:kr}), plays a key role in the $\succeq_{\mathcal{B}}$ test.

\subsection{Bias-based characterization}\label{sect:phase2}

Since several frequent patterns can have the same set of supporting individuals, turning clusters into concepts is crucial in our approach. Biases can be of help. A \emph{bias} concerns anything which constrains the search for theories \cite{UtgoffM82}. In ILP \emph{language bias} has to do with constraints on the clauses in the search space whereas \emph{search bias} has to do with the way a system searches its space of permitted clauses \cite{NedellecRABT96}. The choice criterion for concept intensions has been obtained by combining two orthogonal biases: a language bias and a search bias \cite{LisiE06-ecai}. The former allows the user to define conditions on the form of $\mathcal{O}$-queries to be accepted as concept intensions. E.g., it is possible to state which is the minimum level of description granularity (parameter $minG$) and whether (all) the variables must be ontologically constrained or not. The latter allows the user to define a preference criterion based on $\mathcal{B}$-subsumption. More precisely, it is possible to state whether the \emph{most general description (m.g.d.)} or the \emph{most specific description (m.s.d.)} w.r.t. $\succeq_{\mathcal{B}}$ has to be preferrred. Since $\succeq_{\mathcal{B}}$ is not a total order, it can happen that two patterns $P$ and $Q$, belonging to the same language $\mathcal{L}$, can not be compared w.r.t. $\succeq_\mathcal{B}$. In this case, the m.g.d. (resp. m.s.d) of $P$ and $Q$ is the union (resp. conjunction) of $P$ and $Q$.
\begin{example}\label{ex:incomparable-pat}
The patterns 
{\small 
\begin{tabbing}
    MMMM \= MMM \= MMM \= MMM \= MMM \kill
	  \texttt{q(A)} $\leftarrow$ \texttt{speaks(A,B)}, \texttt{believes(A,C)} \& \texttt{A:MiddleEastCountry}, \texttt{B:ArabicLanguage}
\end{tabbing}
}
\noindent and
{\small 
\begin{tabbing}
    MMMM \= MMM \= MMM \= MMM \= MMM \kill
	  \texttt{q(A)} $\leftarrow$ \texttt{believes(A,B)}, \texttt{speaks(A,C)} \& \texttt{A:MiddleEastCountry}, \texttt{B:MuslimReligion}
\end{tabbing}
}
\noindent have the same answer set \{\texttt{'ARM'}, \texttt{'IR'}\} but are incomparable w.r.t. $\succeq_\mathcal{B}$.
Their m.g.d. is the union of the two:
{\small 
\begin{tabbing}
    MMMM \= MMM \= MMM \= MMM \= MMM \kill
	  \texttt{q(A)} $\leftarrow$ \texttt{speaks(A,B)}, \texttt{believes(A,C)} \& \texttt{A:MiddleEastCountry}, \texttt{B:ArabicLanguage}\\
	  \texttt{q(A)} $\leftarrow$ \texttt{believes(A,B)}, \texttt{speaks(A,C)} \& \texttt{A:MiddleEastCountry}, \texttt{B:MuslimReligion}
\end{tabbing}
}
Their m.s.d. is the conjunction of the two:
{\small 
\begin{tabbing}
    MMMM \= MMM \= MMM \= MMM \= MMM \kill
	  \texttt{q(A)} $\leftarrow$ \texttt{believes(A,B)}, \texttt{speaks(A,C)}, \texttt{speaks(A,D)}, \texttt{believes(A,E)} \&\\
	  \> \texttt{A:MiddleEastCountry}, \texttt{B:MuslimReligion}, \texttt{C:ArabicLanguage}
\end{tabbing}
}
The extension of the subsequent concept will be \{\texttt{'ARM'}, \texttt{'IR'}\}.
\end{example}

The two biases are combined as follows. For each frequent pattern $P \in \mathcal{L}$ that fulfills the language bias specification, the procedure for building the taxonomy $\mathcal{G}$ from the set $\mathcal{F}=\{\mathcal{F}^{l}_{k}\mid 1 \leq l \leq maxG, 1 \leq k \leq maxD\}$ checks whether a concept $\mathcal{C}$ with $ext(\mathcal{C})=answerset(P)$ already exists in $\mathcal{G}$. If one such concept is not retrieved, a new node $\mathcal{C}$ with $int(\mathcal{C})=P$ and $ext(\mathcal{C})=answerset(P)$ is added to $\mathcal{G}$. Note that the insertion of a node can imply the reorganization of $\mathcal{G}$ to keep it compliant with the subset relation on extents. If the node already occurs in $\mathcal{G}$, its intension is updated according to the search bias specification. 

\subsection{Experimental Results}\label{sect:eval}

In order to test the approach we have extended the ILP system $\mathcal{AL}$-QuIn \cite{Lisi06-ppswr}\footnote{$\mathcal{AL}$-QuIn is implemented with Prolog but equipped with a module for pre-processing OWL ontologies in order to enable Semantic Web Mining applications.} with a module for post-processing frequent patterns into concepts. The goal of the experiments is to provide an empirical evidence of the orthogonality of the two biases and of the potential of their combination as choice criterion. The results reported in the following are obtained for the problem introduced in Example \ref{ex:sw-pat} by setting the parameters for the frequent pattern discovery phase as follows: $maxD=5$, $maxG=3$, $minsup^1=20\%$, $minsup^2=13\%$, and $minsup^3=10\%$. Thus each experiment starts from the same set $\mathcal{F}$ of 53 frequent patterns out of 99 candidate patterns. Also all the experiments require the descriptions to have all the variables ontologically constrained but vary as to the user preferences for the minimum level of description granularity ($minG$) and the search bias (m.g.d./m.s.d.).

\begin{figure}[t]
\begin{center}
\setlength{\epsfxsize}{5.5in}
\centerline{\epsfbox{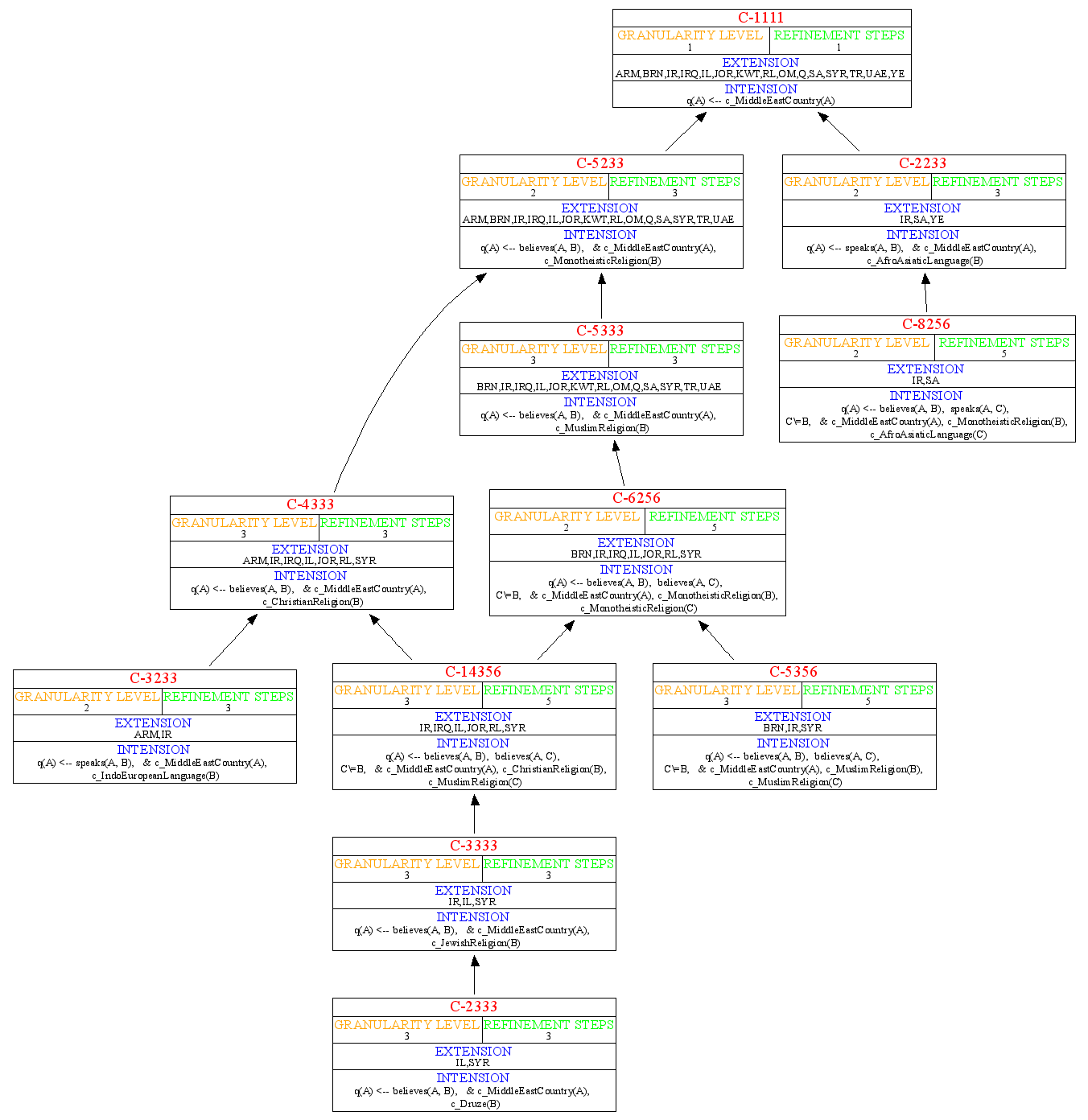}}
\caption{Taxonomy $\mathcal{G}_{\texttt{CIA}}$ obtained with the m.g.d. criterion for $minG=2$.}\label{fig:taxonomy1a}
\end{center}
\end{figure}
The first two experiments both require the descriptions to have all the variables ontologically constrained by concepts from the second granularity level on ($minG=2$). When the m.g.d. criterion is adopted, the procedure of taxonomy building returns the following twelve concepts:
{\small 
\begin{tabbing}
    MMMM \= MMM \= MMM \= MMM \= MMM \kill
	  $\texttt{C-1111} \in \mathcal{F}^1_1$\\
	  \texttt{q(A)} $\leftarrow$ \texttt{A:MiddleEastCountry}\\
	  \{\texttt{ARM}, \texttt{BRN}, \texttt{IR}, \texttt{IRQ}, \texttt{IL}, \texttt{JOR}, \texttt{KWT}, \texttt{RL}, \texttt{OM}, \texttt{Q}, \texttt{SA}, \texttt{SYR}, \texttt{TR}, \texttt{UAE}, \texttt{YE}\}
\end{tabbing}
}
{\small 
\begin{tabbing}
    MMMM \= MMM \= MMM \= MMM \= MMM \kill
	  $\texttt{C-5233} \in \mathcal{F}^2_3$\\
	  \texttt{q(A)} $\leftarrow$ \texttt{believes(A,B)} \& \texttt{A:MiddleEastCountry}, \texttt{B:MonotheisticReligion}\\
	  \{\texttt{ARM}, \texttt{BRN}, \texttt{IR}, \texttt{IRQ}, \texttt{IL}, \texttt{JOR}, \texttt{KWT}, \texttt{RL}, \texttt{OM}, \texttt{Q}, \texttt{SA}, \texttt{SYR}, \texttt{TR}, \texttt{UAE}\}
\end{tabbing}
}
{\small 
\begin{tabbing}
    MMMM \= MMM \= MMM \= MMM \= MMM \kill
	  $\texttt{C-2233} \in \mathcal{F}^2_3$\\
	  \texttt{q(A)} $\leftarrow$ \texttt{speaks(A,B)} \& \texttt{A:MiddleEastCountry}, \texttt{B:AfroAsiaticLanguage}\\
	  \{\texttt{IR}, \texttt{SA}, \texttt{YE}\}
\end{tabbing}
}
{\small 
\begin{tabbing}
    MMMM \= MMM \= MMM \= MMM \= MMM \kill
	  $\texttt{C-3233} \in \mathcal{F}^2_3$\\
	  \texttt{q(A)} $\leftarrow$ \texttt{speaks(A,B)} \& \texttt{A:MiddleEastCountry}, \texttt{B:IndoEuropeanLanguage}\\
	  \{\texttt{ARM}, \texttt{IR}\}
\end{tabbing}
}
{\small 
\begin{tabbing}
    MMMM \= MMM \= MMM \= MMM \= MMM \kill
	  $\texttt{C-8256} \in \mathcal{F}^2_5$\\
	  \texttt{q(A)} $\leftarrow$ \texttt{speaks(A,B)}, \texttt{believes(A,C)} \& \\
	  \> \texttt{A:MiddleEastCountry}, \texttt{B:AfroAsiaticLanguage}, \texttt{C:MonotheisticReligion}\\
	  \{\texttt{IR}, \texttt{SA}\}
\end{tabbing}
}
{\small 
\begin{tabbing}
    MMMM \= MMM \= MMM \= MMM \= MMM \kill
	  $\texttt{C-6256} \in \mathcal{F}^2_5$\\
	  \texttt{q(A)} $\leftarrow$ \texttt{believes(A,B)}, \texttt{believes(A,C)} \& \\
	  \> \texttt{A:MiddleEastCountry}, \texttt{B:MonotheisticReligion}, \texttt{C:MonotheisticReligion}\\
	  \{\texttt{BRN}, \texttt{IR}, \texttt{IRQ}, \texttt{IL}, \texttt{JOR}, \texttt{RL}, \texttt{SYR}\}
\end{tabbing}
}
{\small 
\begin{tabbing}
    MMMM \= MMM \= MMM \= MMM \= MMM \kill
	  $\texttt{C-2333} \in \mathcal{F}^3_3$\\
	  \texttt{q(A)} $\leftarrow$ \texttt{believes(A,'Druze')} \& \texttt{A:MiddleEastCountry}\\
	  \{\texttt{IL}, \texttt{SYR}\}
\end{tabbing}
}
{\small 
\begin{tabbing}
    MMMM \= MMM \= MMM \= MMM \= MMM \kill
	  $\texttt{C-3333} \in \mathcal{F}^3_3$\\
	  \texttt{q(A)} $\leftarrow$ \texttt{believes(A,B)} \& \texttt{A:MiddleEastCountry}, \texttt{B:JewishReligion}\\
	  \{\texttt{IR}, \texttt{IL}, \texttt{SYR}\}
\end{tabbing}
}
{\small 
\begin{tabbing}
    MMMM \= MMM \= MMM \= MMM \= MMM \kill
	  $\texttt{C-4333} \in \mathcal{F}^3_3$\\
	  \texttt{q(A)} $\leftarrow$ \texttt{believes(A,B)} \& \texttt{A:MiddleEastCountry}, \texttt{B:ChristianReligion}\\
	  \{\texttt{ARM}, \texttt{IR}, \texttt{IRQ}, \texttt{IL}, \texttt{JOR}, \texttt{RL}, \texttt{SYR}\}
\end{tabbing}
}
{\small 
\begin{tabbing}
    MMMM \= MMM \= MMM \= MMM \= MMM \kill
	  $\texttt{C-5333} \in \mathcal{F}^3_3$\\
	  \texttt{q(A)} $\leftarrow$ \texttt{believes(A,B)} \& \texttt{A:MiddleEastCountry}, \texttt{B:MuslimReligion}\\
	  \{\texttt{BRN}, \texttt{IR}, \texttt{IRQ}, \texttt{IL}, \texttt{JOR}, \texttt{KWT}, \texttt{RL}, \texttt{OM}, \texttt{Q}, \texttt{SA}, \texttt{SYR}, \texttt{TR}, \texttt{UAE}\}
\end{tabbing}
}
{\small 
\begin{tabbing}
    MMMM \= MMM \= MMM \= MMM \= MMM \kill
	  $\texttt{C-14356} \in \mathcal{F}^3_5$\\
	  \texttt{q(A)} $\leftarrow$ \texttt{believes(A,B)}, \texttt{believes(A,C)} \& \\
	  \> \texttt{A:MiddleEastCountry}, \texttt{B:ChristianReligion}, \texttt{C:MuslimReligion}\\
	  \{\texttt{IR}, \texttt{IRQ}, \texttt{IL}, \texttt{JOR}, \texttt{RL}, \texttt{SYR}\}
\end{tabbing}
}
{\small 
\begin{tabbing}
    MMMM \= MMM \= MMM \= MMM \= MMM \kill
	  $\texttt{C-5356} \in \mathcal{F}^3_5$\\
	  \texttt{q(A)} $\leftarrow$ \texttt{believes(A,B)}, \texttt{believes(A,C)} \& \\
	  \> \texttt{A:MiddleEastCountry}, \texttt{B:MuslimReligion}, \texttt{C:MuslimReligion}\\
	  \{\texttt{BRN}, \texttt{IR}, \texttt{SYR}\}
\end{tabbing}
}
\noindent organized in the DAG $\mathcal{G}_{\texttt{CIA}}$ (see Figure \ref{fig:taxonomy1a}). They are numbered according to the chronological order of insertion in $\mathcal{G}_{\texttt{CIA}}$ and annotated with information of the generation step. From a qualitative point of view, concepts $\texttt{C-2233}$\footnote{$\texttt{C-2233}$ is less populated than expected because $\mathcal{B}_{\texttt{CIA}}$ does not provide facts on the languages spoken for all countries.} and $\texttt{C-5333}$ well characterize Middle East countries. Armenia (\texttt{ARM}), as opposite to Iran (\texttt{IR}), does not fall in these concepts. It rather belongs to the weaker characterizations $\texttt{C-3233}$ and $\texttt{C-4333}$. This suggests that our procedure performs a 'sensible' clustering. Indeed Armenia is a well-known borderline case for the geo-political concept of Middle East, though the Armenian is usually listed among Middle Eastern ethnic groups. Modern experts tend nowadays to consider it as part of Europe, therefore out of Middle East. But in 1996 the on-line CIA World Fact Book still considered Armenia as part of Asia.

\begin{figure}[t]
\begin{center}
\setlength{\epsfxsize}{5.5in}
\centerline{\epsfbox{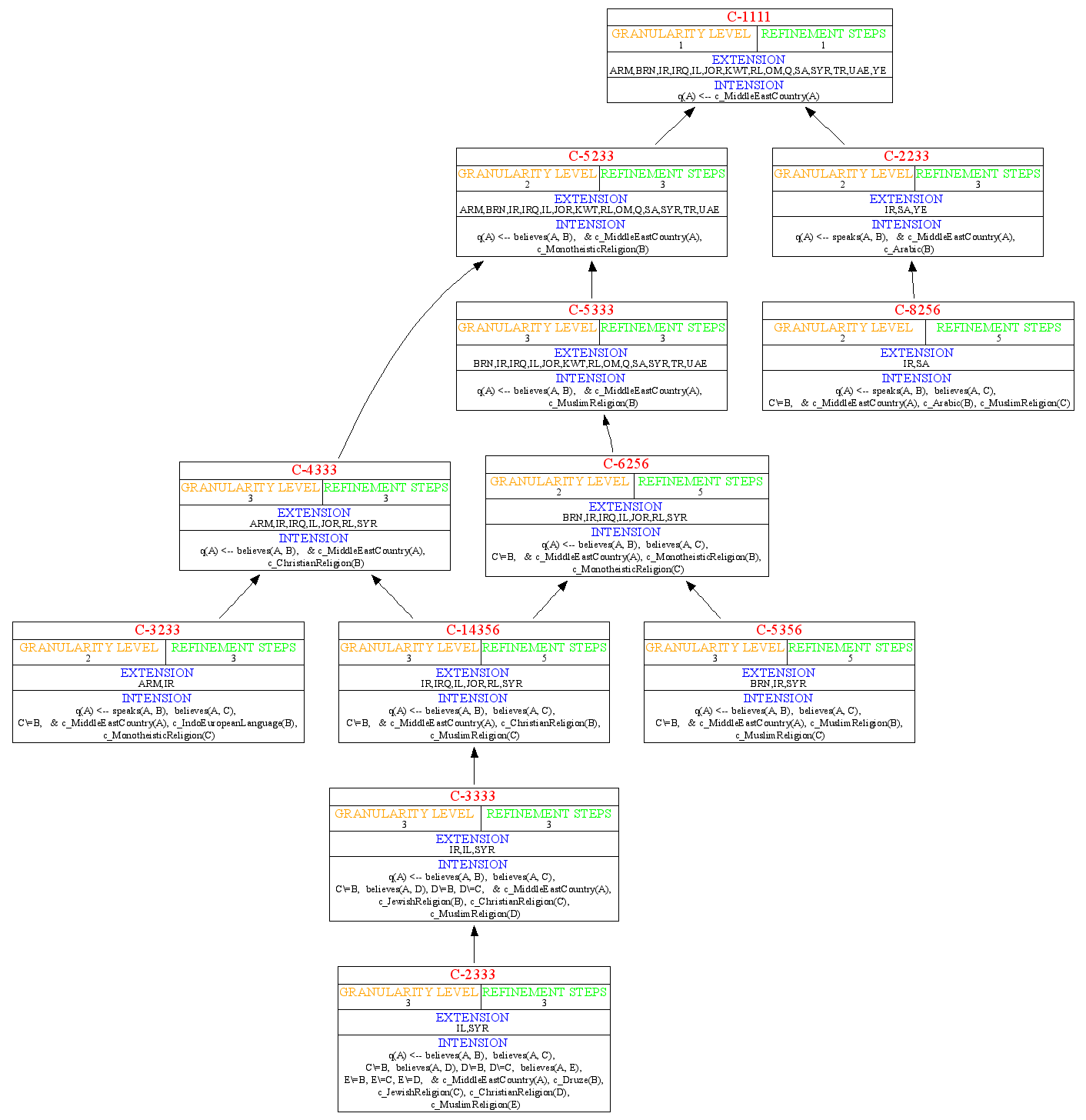}}
\caption{Taxonomy $\mathcal{G}_{\texttt{CIA}}^{\prime}$ obtained with the m.s.d. criterion for $minG=2$.}\label{fig:taxonomy1b}
\end{center}
\end{figure}
When the m.s.d. criterion is adopted (see Figure \ref{fig:taxonomy1b}), the intensions for the concepts $\texttt{C-2233}$, $\texttt{C-3233}$, $\texttt{C-8256}$, $\texttt{C-2333}$ and $\texttt{C-3333}$ change as follows:
{\small 
\begin{tabbing}
    MMMM \= MMM \= MMM \= MMM \= MMM \kill
	  $\texttt{C-2233} \in \mathcal{F}^2_3$\\
	  \texttt{q(A)} $\leftarrow$ \texttt{speaks(A,B)} \& \texttt{A:MiddleEastCountry}, \texttt{B:ArabicLanguage}\\
	  \{\texttt{IR}, \texttt{SA}, \texttt{YE}\}
\end{tabbing}
}
{\small 
\begin{tabbing}
    MMMM \= MMM \= MMM \= MMM \= MMM \kill
	  $\texttt{C-3233} \in \mathcal{F}^2_3$\\
	  \texttt{q(A)} $\leftarrow$ \texttt{speaks(A,B)} \& \texttt{A:MiddleEastCountry}, \texttt{B:IndoIranianLanguage}\\
	  \{\texttt{ARM}, \texttt{IR}\}
\end{tabbing}
}
{\small 
\begin{tabbing}
    MMMM \= MMM \= MMM \= MMM \= MMM \kill
	  $\texttt{C-8256} \in \mathcal{F}^2_5$\\
	  \texttt{q(A)} $\leftarrow$ \texttt{speaks(A,B)}, \texttt{believes(A,C)} \& \\
	  \> \texttt{A:MiddleEastCountry}, \texttt{B:ArabicLanguage}, \texttt{C:MuslimReligion}\\
	  \{\texttt{IR}, \texttt{SA}\}
\end{tabbing}
}
{\small 
\begin{tabbing}
    MMMM \= MMM \= MMM \= MMM \= MMM \kill
	  $\texttt{C-2333} \in \mathcal{F}^3_3$\\
	  \texttt{q(A)} $\leftarrow$ \texttt{believes(A,'Druze')}, \texttt{believes(A,B)}, \texttt{believes(A,C)}, \texttt{believes(A,D)} \& \\
	  \> \texttt{A:MiddleEastCountry}, \texttt{B:JewishReligion},\\
	  \> \texttt{C:ChristianReligion}, \texttt{D:MuslimReligion}\\
	  \{\texttt{IL}, \texttt{SYR}\}
\end{tabbing}
}
{\small 
\begin{tabbing}
    MMMM \= MMM \= MMM \= MMM \= MMM \kill
	  $\texttt{C-3333} \in \mathcal{F}^3_3$\\
	  \texttt{q(A)} $\leftarrow$ \texttt{believes(A,B)}, \texttt{believes(A,C)}, \texttt{believes(A,D)} \& \\
	  \> \texttt{A:MiddleEastCountry}, \texttt{B:JewishReligion},\\
	  \> \texttt{C:ChristianReligion}, \texttt{D:MuslimReligion}\\
	  \{\texttt{IR}, \texttt{IL}, \texttt{SYR}\}
\end{tabbing}
}
\noindent In particular $\texttt{C-2333}$ and $\texttt{C-3333}$ look quite overfitted to data. Yet overfitting allows us to realize that what distinguishes Israel (\texttt{IL}) and Syria (\texttt{SYR}) from Iran is just the presence of Druze people. Note that the clusters do not change because the search bias only affects the characterization step.

\begin{figure}[t]
\begin{center}
\setlength{\epsfxsize}{5.5in}
\centerline{\epsfbox{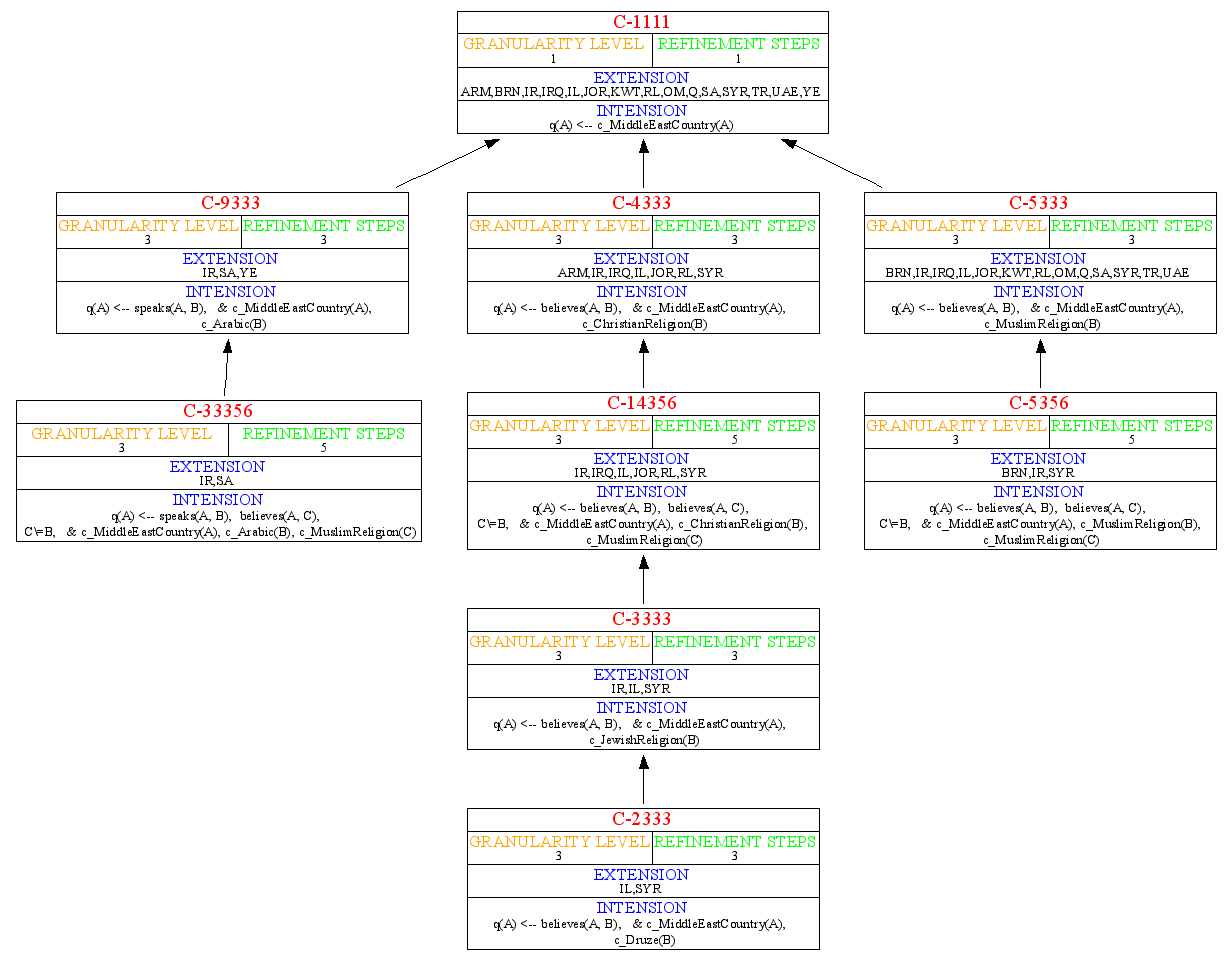}}
\caption{Taxonomy $\mathcal{G}_{\texttt{CIA}}^{\prime\prime}$ obtained with the m.g.d. criterion for $minG=3$.}\label{fig:taxonomy2a}
\end{center}
\end{figure}
The other two experiments further restrict the conditions of the language bias specification. Here only descriptions with variables constrained by concepts of granularity from the third level on ($minG=3$) are considered. When the m.g.d. option is selected, the procedure for taxonomy building returns the following nine concepts: 
{\small 
\begin{tabbing}
    MMMM \= MMM \= MMM \= MMM \= MMM \kill
	  $\texttt{C-1111} \in \mathcal{F}^1_1$\\
	  \texttt{q(A)} $\leftarrow$ \texttt{A:MiddleEastCountry}\\
	  \{\texttt{ARM}, \texttt{BRN}, \texttt{IR}, \texttt{IRQ}, \texttt{IL}, \texttt{JOR}, \texttt{KWT}, \texttt{RL}, \texttt{OM}, \texttt{Q}, \texttt{SA}, \texttt{SYR}, \texttt{TR}, \texttt{UAE}, \texttt{YE}\}
\end{tabbing}
}
{\small 
\begin{tabbing}
    MMMM \= MMM \= MMM \= MMM \= MMM \kill
	  $\texttt{C-9333} \in \mathcal{F}^3_3$\\
	  \texttt{q(A)} $\leftarrow$ \texttt{speaks(A,B)} \& \texttt{A:MiddleEastCountry}, \texttt{B:ArabicLanguage}\\
	  \{\texttt{IR}, \texttt{SA}, \texttt{YE}\}
\end{tabbing}
}
{\small 
\begin{tabbing}
    MMMM \= MMM \= MMM \= MMM \= MMM \kill
	  $\texttt{C-2333} \in \mathcal{F}^3_3$\\
	  \texttt{q(A)} $\leftarrow$ \texttt{believes(A,'Druze')} \& \texttt{A:MiddleEastCountry}\\
	  \{\texttt{IL}, \texttt{SYR}\}
\end{tabbing}
}
{\small 
\begin{tabbing}
    MMMM \= MMM \= MMM \= MMM \= MMM \kill
	  $\texttt{C-3333} \in \mathcal{F}^3_3$\\
	  \texttt{q(A)} $\leftarrow$ \texttt{believes(A,B)} \& \texttt{A:MiddleEastCountry}, \texttt{B:JewishReligion}\\
	  \{\texttt{IR}, \texttt{IL}, \texttt{SYR}\}
\end{tabbing}
}
{\small 
\begin{tabbing}
    MMMM \= MMM \= MMM \= MMM \= MMM \kill
	  $\texttt{C-4333} \in \mathcal{F}^3_3$\\
	  \texttt{q(A)} $\leftarrow$ \texttt{believes(A,B)} \& \texttt{A:MiddleEastCountry}, \texttt{B:ChristianReligion}\\
	  \{\texttt{ARM}, \texttt{IR}, \texttt{IRQ}, \texttt{IL}, \texttt{JOR}, \texttt{RL}, \texttt{SYR}\}
\end{tabbing}
}
{\small 
\begin{tabbing}
    MMMM \= MMM \= MMM \= MMM \= MMM \kill
	  $\texttt{C-5333} \in \mathcal{F}^3_3$\\
	  \texttt{q(A)} $\leftarrow$ \texttt{believes(A,B)} \& \texttt{A:MiddleEastCountry}, \texttt{B:MuslimReligion}\\
	  \{\texttt{BRN}, \texttt{IR}, \texttt{IRQ}, \texttt{IL}, \texttt{JOR}, \texttt{KWT}, \texttt{RL}, \texttt{OM}, \texttt{Q}, \texttt{SA}, \texttt{SYR}, \texttt{TR}, \texttt{UAE}\}
\end{tabbing}
}
{\small 
\begin{tabbing}
    MMMM \= MMM \= MMM \= MMM \= MMM \kill
	  $\texttt{C-33356} \in \mathcal{F}^3_5$\\
	  \texttt{q(A)} $\leftarrow$ \texttt{speaks(A,B)}, \texttt{believes(A,C)} \& \\
	  \> \texttt{A:MiddleEastCountry}, \texttt{B:ArabicLanguage}, \texttt{C:MuslimReligion}\\
	  \{\texttt{IR}, \texttt{SA}\}
\end{tabbing}
}
{\small 
\begin{tabbing}
    MMMM \= MMM \= MMM \= MMM \= MMM \kill
	  $\texttt{C-14356} \in \mathcal{F}^3_5$\\
	  \texttt{q(A)} $\leftarrow$ \texttt{believes(A,B)}, \texttt{believes(A,C)} \& \\
	  \> \texttt{A:MiddleEastCountry}, \texttt{B:ChristianReligion}, \texttt{C:MuslimReligion}\\
	  \{\texttt{IR}, \texttt{IRQ}, \texttt{IL}, \texttt{JOR}, \texttt{RL}, \texttt{SYR}\}
\end{tabbing}
}
{\small 
\begin{tabbing}
    MMMM \= MMM \= MMM \= MMM \= MMM \kill
	  $\texttt{C-5356} \in \mathcal{F}^3_5$\\
	  \texttt{q(A)} $\leftarrow$ \texttt{believes(A,B)}, \texttt{believes(A,C)} \& \\
	  \> \texttt{A:MiddleEastCountry}, \texttt{B:MuslimReligion}, \texttt{C:MuslimReligion}\\
	  \{\texttt{BRN}, \texttt{IR}, \texttt{SYR}\}
\end{tabbing}
}
\noindent organized in a DAG $\mathcal{G}_{\texttt{CIA}}^{\prime\prime}$ (see Figure \ref{fig:taxonomy2a}) which partially reproduces $\mathcal{G}_{\texttt{CIA}}^{\prime}$. Note that the stricter conditions set in the language bias cause three concepts occurring in $\mathcal{G}_{\texttt{CIA}}^{\prime}$ not to appear in $\mathcal{G}_{\texttt{CIA}}^{\prime\prime}$: the scarsely significant $\texttt{C-5233}$ and $\texttt{C-6256}$, and the quite interesting $\texttt{C-3233}$. Therefore the language bias can prune the space of clusters. Note that the other concepts of $\mathcal{G}_{\texttt{CIA}}^{\prime}$ emerged at $l=2$ do remain in $\mathcal{G}_{\texttt{CIA}}^{\prime\prime}$ as clusters but with a different characterization: $\texttt{C-9333}$ and $\texttt{C-33356}$ instead of $\texttt{C-2233}$ and $\texttt{C-8256}$, respectively.

\begin{figure}[t]
\begin{center}
\setlength{\epsfxsize}{5.5in}
\centerline{\epsfbox{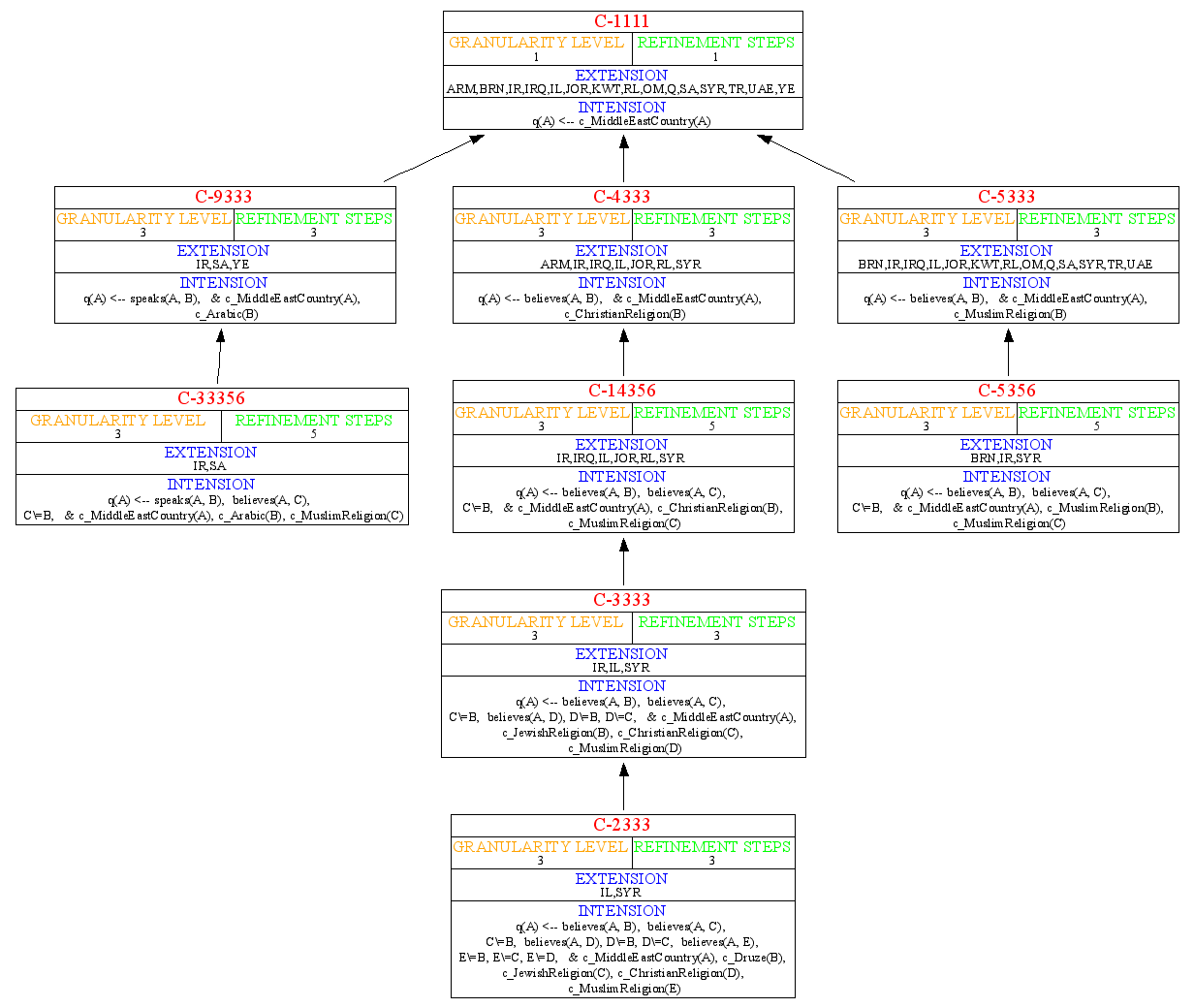}}
\caption{Taxonomy $\mathcal{G}_{\texttt{CIA}}^{\prime\prime\prime}$ obtained with the m.s.d. criterion for $minG=3$.}\label{fig:taxonomy2b}
\end{center}
\end{figure}
When the m.s.d. condition is chosen (see Figure \ref{fig:taxonomy2b}), the intensions for the concepts $\texttt{C-2333}$ and $\texttt{C-3333}$ change analogously to $\mathcal{G}_{\texttt{CIA}}^{\prime}$. Note that both $\mathcal{G}_{\texttt{CIA}}^{\prime\prime}$ and $\mathcal{G}_{\texttt{CIA}}^{\prime\prime\prime}$ are hierarchical taxonomies. It can be empirically observed that the possibility of producing a hierarchy increases as the conditions of the language bias become stricter.

\section{Conclusions}\label{sect:concl}

Building rules on top of ontologies for the Semantic Web is a task that can be automated by applying Machine Learning algorithms to data expressed with hybrid formalisms combining DLs and Horn clauses.
Learning in DL-based hybrid languages has very recently attracted attention in the ILP community. In \cite{RouveirolV2000} the
chosen language is \textsc{Carin}-$\mathcal{ALN}$, therefore example coverage and
subsumption between two hypotheses are based on the existential entailment algorithm
of \textsc{Carin} \cite{LevyR98}. Following \cite{RouveirolV2000}, Kietz studies the learnability of
\textsc{Carin}-$\mathcal{ALN}$, thus providing a pre-processing method which enables
ILP systems to learn \textsc{Carin}-$\mathcal{ALN}$ rules \cite{Kietz03}. Closely
related to DL-based hybrid systems are the proposals arising from the study of
many-sorted logics, where a first-order language is combined with a sort language
which can be regarded as an elementary DL \cite{Frisch91}. In this respect the study
of a sorted downward refinement \cite{Frisch99} can be also considered a contribution to
learning in hybrid languages. In this paper we have proposed a general framework for learning in $\mathcal{AL}$-log. We would like to emphasize that the \emph{DL-safeness} and the \emph{decidability} of $\mathcal{AL}$-log are two desirable properties which are particularly appreciated both in ILP and in the Semantic Web application area.

As an instantiation of the framework we have considered the case of characteristic induction from interpretations, more precisely the task of Frequent Pattern Discovery, and an application to Ontology Refinement. The specific problem at hand takes an ontology as input and returns subconcepts of one of the concepts in the ontology. A distinguishing feature of our setting for this problem is that the intensions of these subconcepts are in the form of rules that are automatically built by discovering strong associations between concepts in the input ontology. The idea of resorting to Frequent Pattern Discovery in Ontology Learning has been already investigated in \cite{MaedcheS00}. Yet there are several differences between \cite{MaedcheS00} and the present work: \cite{MaedcheS00} is conceived for Ontology Extraction instead of Ontology Refinement, uses generalized association patterns (bottom-up search) instead of multi-level association patterns (top-down search), adopts propositional logic instead of FOL. Within the same application area, \cite{MaedcheZ02} proposes a distance-based method for clustering in RDF which is not conceptual. 
Also the relation between Frequent Pattern Discovery and Concept Formation as such has never been investigated. Rather our pattern-based approach to clustering is inspired by \cite{XiongSRK05}. Some contact points can be also found with \cite{ZimmermannDR04} that defines the problem of cluster-grouping and a solution to it that integrates Subgroup Discovery, Correlated Pattern Mining and Conceptual Clustering. Note that neither \cite{XiongSRK05} nor \cite{ZimmermannDR04} deal with (fragments of) FOL. Conversely, \cite{Stumme04} combines the notions of frequent \textsc{Datalog} query and iceberg concept lattices to upgrade Formal Concept Analysis (a well-established and widely used approach for Conceptual Clustering) to FOL. Generally speaking, very few works on Conceptual Clustering and Concept Formation in FOL can be found in the literature. They vary as for the approaches (distance-based, probabilistic, etc.) and/or the representations (description logics, conceptual graphs, E/R models, etc.) adopted.
The closest work to ours is Vrain's proposal \cite{Vrain96} of a top-down incremental but distance-based method for Conceptual Clustering in a mixed object-logical representation. 

For the future we plan to extensively evaluate this approach on significantly big and expressive ontologies. Without doubt, there is a lack of evaluation standards in Ontology Learning. Comparative work in this field would help an ontology engineer to choose the appropriate method. One step in this direction is the framework presented in \cite{BissonNC00} but it is conceived for Ontology Extraction. The evaluation of our approach can follow the criteria outlined in \cite{DellschaftS06} or criteria from the ML tradition like measuring the cluster validity \cite{HalkidiBV01}, or the category utility \cite{Fisher87}. Anyway, due to the peculiarities of our approach, the evaluation itself requires a preliminary work from the methodological point of view. Regardless of performance, each approach has its own benefits. Our approach has the advantages of dealing with expressive ontologies and being conceptual. One such approach, and in particular its ability of forming concepts with an intensional description in the form of rule, can support many of the use cases defined by the \emph{W3C Rule Interchange Format Working Group}. Another direction of future work can be the extension of the present work towards hybrid formalisms, e.g. \cite{MotikSS04}, that are more expressive than $\mathcal{AL}$-log and more inspiring for prototipical SWRL reasoners. Also we would like to investigate other instantiations of the framework, e.g. the ones in the case of discriminant induction to learn predictive rules.




\appendix
\section{The semantic mark-up language OWL}\label{sect:owl}

The Web Ontology Language OWL is a semantic mark-up language for publishing and sharing ontologies on the World Wide Web \cite{HorrocksP-SvH03}. An OWL ontology is an RDF graph, which is in turn a set of RDF triples. As with any RDF graph, an OWL ontology graph can be written in many different syntactic forms. However, the meaning of an OWL ontology is solely determined by the RDF graph. Thus, it is allowable to use other syntactic RDF/XML forms, as long as these result in the same underlying set of RDF triples.

OWL provides three increasingly expressive sublanguages designed for use by specific communities of implementers and users.
\begin{itemize}
	\item \textit{OWL Lite} supports those users primarily needing a classification hierarchy and simple constraints. E.g., while it supports cardinality constraints, it only permits cardinality values of 0 or 1. It should be simpler to provide tool support for OWL Lite than its more expressive relatives, and OWL Lite provides a quick migration path for thesauri and other taxonomies. OWL Lite also has a lower formal complexity than OWL DL.
	\item \textit{OWL DL} supports those users who want the maximum expressiveness while retaining computational completeness and decidability. OWL DL includes all OWL language constructs, but they can be used only under certain restrictions (e.g., while a class may be a subclass of many classes, a class cannot be an instance of another class). OWL DL is so named due to its correspondence with the very expressive DL $\mathcal{SHOIN}(\textbf{D})$ \cite{HorrocksST00} which thus provides a logical foundation to OWL. The mapping from $\mathcal{ALC}$ to OWL is reported in Table \ref{tab:ALCvsOWL}.
	\item \textit{OWL Full} is meant for users who want maximum expressiveness and the syntactic freedom of RDF with no computational guarantees. For example, in OWL Full a class can be treated simultaneously as a collection of individuals and as an individual in its own right. OWL Full allows an ontology to augment the meaning of the pre-defined (RDF or OWL) vocabulary. It is unlikely that any reasoning software will be able to support complete reasoning for every feature of OWL Full.
\end{itemize}
Each of these sublanguages is an extension of its simpler predecessor, both in what can be legally expressed and in what can be validly concluded.

\begin{table}[ht]
\small
\caption{Mapping from $\mathcal{ALC}$ to OWL}	\label{tab:ALCvsOWL}
\begin{center}
\begin{tabular}{l|l}
\hline
$\neg C$ & \texttt{<owl:Class>}\\
         & \hskip 0.3cm \texttt{<owl:complementOf>}\texttt{<owl:Class rdf:ID="C" />}\texttt{</owl:complementOf>}\\
         & \texttt{</owl:Class>}\\
$C \sqcap D$ & \texttt{<owl:Class>}\\
             & \hskip 0.3cm \texttt{<owl:intersectionOf rdf:parseType="Collection">}\\
             & \hskip 0.6cm \texttt{<owl:Class rdf:ID="C" />}\texttt{<owl:Class rdf:ID="D" />}\\
             & \hskip 0.3cm \texttt{</owl:intersectionOf>}\\
             & \texttt{</owl:Class>}\\
$C \sqcup D$ & \texttt{<owl:Class>}\\
             & \hskip 0.3cm \texttt{<owl:unionOf rdf:parseType="Collection">}\\
             & \hskip 0.6cm \texttt{<owl:Class rdf:ID="C" />}\texttt{<owl:Class rdf:ID="D" />}\\
             & \hskip 0.3cm \texttt{</owl:unionOf>}\\
             & \texttt{</owl:Class>}\\
$\exists R.C$ & \texttt{<owl:Restriction>}\\
              & \hskip 0.3cm \texttt{<owl:onProperty rdf:resource="\#R" />}\\
              & \hskip 0.3cm \texttt{<owl:someValuesFrom rdf:resource="\#C" />}\\
              & \texttt{</owl:Restriction>}\\
$\forall R.C$ & \texttt{<owl:Restriction>}\\
              & \hskip 0.3cm \texttt{<owl:onProperty rdf:resource="\#R" />}\\
              & \hskip 0.3cm \texttt{<owl:allValuesFrom rdf:resource="\#C" />}\\
              & \texttt{</owl:Restriction>}\\
\hline
$C \equiv D$ & \texttt{<owl:Class rdf:ID="C">}\\
             & \hskip 0.3cm \texttt{<owl:sameAs rdf:resource="\#D" />}\\
             & \texttt{</owl:Class>}\\
$C \sqsubseteq D$ & \texttt{<owl:Class rdf:ID="C">}\\
                  & \hskip 0.3cm \texttt{<rdfs:subClassOf rdf:resource="\#D" />}\\
                  & \texttt{</owl:Class>}\\
\hline
$a:C$ & \texttt{<C rdf:ID="a" />}\\
$\langle a,b \rangle:R$ & \texttt{<C rdf:ID="a">}\texttt{<R rdf:resource="\#b" />}\\
\hline
\end{tabular}
\end{center}
\end{table}


\begin{thebibliography}{}

\bibitem[\protect\citeauthoryear{Butcher}{Butcher}{1981}]{Butcher}
{\sc Butcher, J.} 1981.
\newblock {\em Copy-editing: The Cambridge Handbook}.
\newblock Cambridge University Press.

\bibitem[\protect\citeauthoryear{{{C}adence {R}esearch {S}ystems}}{{{C}adence
  {R}esearch {S}ystems}}{1994}]{crs:chez}
{\sc {{C}adence {R}esearch {S}ystems}}. 1994.
\newblock {\it {C}hez} {S}cheme {R}eference {M}anual.

\bibitem[\protect\citeauthoryear{Cameron and Ito}{Cameron and
  Ito}{1984}]{ci:gramps}
{\sc Cameron, R.~D.} {\sc and} {\sc Ito, M.~R.} 1984.
\newblock Grammar-based definition of metaprogramming systems.
\newblock {\em {ACM} Transactions on Programming Languages and Systems\/}~{\em
  6,\/}~1 (Jan.), 20--54.

\bibitem[\protect\citeauthoryear{Grossman}{Grossman}{1982}]{Chicago}
{\sc Grossman, J.}, Ed. 1982.
\newblock {\em The Chicago Manual of Style}.
\newblock University of Chicago Press.

\bibitem[\protect\citeauthoryear{Lamport}{Lamport}{1986}]{LaTeX}
{\sc Lamport, L.} 1986.
\newblock {\em \LaTeX: A Document Preparation System\/}, 2 ed.
\newblock Addison-Wesley, New York.

\end{thebibliography}


\begin{thebibliography}{}

\bibitem[\protect\citeauthoryear{Apt and Bol}{Apt and Bol}{1994}]{apt:94}
{\sc Apt, K.} {\sc and} {\sc Bol, R.} 1994.
\newblock Logic programming and negation: A survey.
\newblock {\em Journal of Logic Programming\/}~{\em 19,20}, 9--71.

\bibitem[\protect\citeauthoryear{Brewka}{Brewka}{1996}]{brewka:plp}
{\sc Brewka, G.} 1996.
\newblock Well-founded semantics for extended logic programs with dynamic
  preferences.
\newblock {\em Journal of Artificial Intelligence Research\/}~{\em 4}, 19--36.

\bibitem[\protect\citeauthoryear{Brewka and Eiter}{Brewka and
  Eiter}{1999}]{be:aij}
{\sc Brewka, G.} {\sc and} {\sc Eiter, T.} 1999.
\newblock Preferred answer sets for extended logic programs.
\newblock {\em Artificial Intelligence\/}~{\em 109}, 297--356.

\bibitem[\protect\citeauthoryear{Choelwinski}{Choelwinski}{1994}]{ch:stra}
{\sc Choelwinski, P.} 1994.
\newblock Stratified default logic.
\newblock In {\em Proceedings of Computer Science Logic}. Springer, LNCS 933,
  456--470.

\bibitem[\protect\citeauthoryear{Das}{Das}{1992}]{lp:92}
{\sc Das, S.} 1992.
\newblock {\em Deductive Database and Logic Programming}.
\newblock Addison-Wesley Publishers Ltd.

\bibitem[\protect\citeauthoryear{Delgrande, Schaub, and Tompits}{Delgrande
  et~al\mbox{.}}{2000}]{d:plp}
{\sc Delgrande, J.}, {\sc Schaub, T.}, {\sc and} {\sc Tompits, H.} 2000.
\newblock Logic programs with complied preferences.
\newblock In {\em Proceedings of the 14th European Conference on Artificial
  Intelligence (ECAI-2000)}. 392--398.

\bibitem[\protect\citeauthoryear{Gelfond and Lifschitz}{Gelfond and
  Lifschitz}{1988}]{gl:stable}
{\sc Gelfond, M.} {\sc and} {\sc Lifschitz, V.} 1988.
\newblock The stable model semantics for logic programming.
\newblock In {\em Proceedings of the Fifth Joint International Conference and
  Symposium}. MIT Press, 1070--1080.

\bibitem[\protect\citeauthoryear{Gelfond and Lifschitz}{Gelfond and
  Lifschitz}{1991}]{gl:logic}
{\sc Gelfond, M.} {\sc and} {\sc Lifschitz, V.} 1991.
\newblock Classical negation in logic programs and disjunctive databases.
\newblock {\em New Generation Computing\/}~{\em 9}, 365--386.

\bibitem[\protect\citeauthoryear{Gelfond and Son}{Gelfond and
  Son}{1998}]{ms:98}
{\sc Gelfond, M.} {\sc and} {\sc Son, T.} 1998.
\newblock Reasoning with prioritized defaults.
\newblock In {\em LNAI 1471}. Springer, 164--224.

\bibitem[\protect\citeauthoryear{Grosof}{Grosof}{1997}]{g:97}
{\sc Grosof, B.} 1997.
\newblock Prioritized conflict handling for logic programs.
\newblock In {\em Proceedings of the 1997 International Logic Program Symposium
  (ILPS'97)}. MIT Press, 197--212.

\bibitem[\protect\citeauthoryear{Lifschitz and Turner}{Lifschitz and
  Turner}{1994}]{lt:lp94}
{\sc Lifschitz, V.} {\sc and} {\sc Turner, H.} 1994.
\newblock Splitting a logic program.
\newblock In {\em Proceedings of Eleventh International Conference on Logic
  Programming}. MIT Press, 23--37.

\bibitem[\protect\citeauthoryear{Schaub and Wang}{Schaub and
  Wang}{2001}]{sw:01}
{\sc Schaub, T.} {\sc and} {\sc Wang, K.} 2001.
\newblock A comparative study of logic programs with preference.
\newblock In {\em Proceedings of the 17th International Joint Conference on
  Artificial Intelligence (IJCAI-2001)}. Morgan Kaufmann Publishers Inc.,
  597--602.

\bibitem[\protect\citeauthoryear{Wang, Zhou, and Lin}{Wang
  et~al\mbox{.}}{2000}]{w:plp}
{\sc Wang, K.}, {\sc Zhou, L.}, {\sc and} {\sc Lin, F.} 2000.
\newblock Alternating fixpoint theory for logic programs with priority.
\newblock In {\em Proceedings of International Joint Conference on
  Computational Logic (CL-2000)}. 164--178.

\bibitem[\protect\citeauthoryear{Zhang}{Zhang}{1999}]{yan:iclp99}
{\sc Zhang, Y.} 1999.
\newblock Monotonicity in rule based update.
\newblock In {\em Proceedings of the 1999 International Conference on Logic
  Programming (ICLP'99)}. MIT Press, 471--485.

\bibitem[\protect\citeauthoryear{Zhang}{Zhang}{2001}]{yan:00lp}
{\sc Zhang, Y.} 2001.
\newblock The complexity of logic program update.
\newblock In {\em Proceedings of the 14th Australian Joint Conference on
  Artificial Intelligence (AI2001)}. Springer, LNAI 2256, 630--643.

\bibitem[\protect\citeauthoryear{Zhang and Foo}{Zhang and Foo}{1997}]{yan:plp}
{\sc Zhang, Y.} {\sc and} {\sc Foo, N.} 1997.
\newblock Answer sets for prioritized logic programs.
\newblock In {\em Proceedings of the 1997 International Logic Programming
  Symposium (ILPS'97)}. MIT Press, 69--83.

\end{thebibliography}


\begin{thebibliography}{}

\bibitem[\protect\citeauthoryear{Baader, Calvanese, McGuinness, Nardi, and
  Patel-Schneider}{Baader et~al\mbox{.}}{2003}]{BaaderCMcGNPS03}
{\sc Baader, F.}, {\sc Calvanese, D.}, {\sc McGuinness, D.}, {\sc Nardi, D.},
  {\sc and} {\sc Patel-Schneider, P.}, Eds. 2003.
\newblock {\em The Description Logic Handbook: Theory, Implementation and
  Applications}.
\newblock Cambridge University Press.

\bibitem[\protect\citeauthoryear{Berners-Lee, Hendler, and Lassila}{Berners-Lee
  et~al\mbox{.}}{2001}]{Berners-Lee01}
{\sc Berners-Lee, T.}, {\sc Hendler, J.}, {\sc and} {\sc Lassila, O.} 2001.
\newblock The {S}emantic {W}eb.
\newblock {\em Scientific American\/}~{\em May}.

\bibitem[\protect\citeauthoryear{Bisson, Nedellec, and Ca{\~n}amero}{Bisson
  et~al\mbox{.}}{2000}]{BissonNC00}
{\sc Bisson, G.}, {\sc Nedellec, C.}, {\sc and} {\sc Ca{\~n}amero, D.} 2000.
\newblock Designing clustering methods for ontology building - the {M}o'{K}
  workbench.
\newblock In {\em ECAI Workshop on Ontology Learning}, {S.~Staab},
  {A.~Maedche}, {C.~Nedellec}, {and} {P.~Wiemer-Hastings}, Eds. CEUR Workshop
  Proceedings, vol.~31. CEUR-WS.org.

\bibitem[\protect\citeauthoryear{Blockeel, De~Raedt, Jacobs, and
  Demoen}{Blockeel et~al\mbox{.}}{1999}]{Blockeel99}
{\sc Blockeel, H.}, {\sc De~Raedt, L.}, {\sc Jacobs, N.}, {\sc and} {\sc
  Demoen, B.} 1999.
\newblock Scaling {U}p {I}nductive {L}ogic {P}rogramming by {L}earning from
  {I}nterpretations.
\newblock {\em Data Mining and Knowledge Discovery\/}~{\em 3}, 59--93.

\bibitem[\protect\citeauthoryear{Borgida}{Borgida}{1996}]{Borgida96}
{\sc Borgida, A.} 1996.
\newblock On the relative expressiveness of description logics and predicate
  logics.
\newblock {\em Artificial Intelligence\/}~{\em 82,\/}~1--2, 353--367.

\bibitem[\protect\citeauthoryear{Buntine}{Buntine}{1988}]{Buntine88}
{\sc Buntine, W.} 1988.
\newblock Generalized subsumption and its application to induction and
  redundancy.
\newblock {\em Artificial Intelligence\/}~{\em 36,\/}~2, 149--176.

\bibitem[\protect\citeauthoryear{Ceri, Gottlob, and Tanca}{Ceri
  et~al\mbox{.}}{1990}]{Ceri90}
{\sc Ceri, S.}, {\sc Gottlob, G.}, {\sc and} {\sc Tanca, L.} 1990.
\newblock {\em Logic Programming and Databases}.
\newblock Springer.

\bibitem[\protect\citeauthoryear{De~Raedt}{De~Raedt}{1997}]{DeRaedt97}
{\sc De~Raedt, L.} 1997.
\newblock Logical {S}ettings for {C}oncept-{L}earning.
\newblock {\em Artificial Intelligence\/}~{\em 95,\/}~1, 187--201.

\bibitem[\protect\citeauthoryear{De~Raedt and Dehaspe}{De~Raedt and
  Dehaspe}{1997}]{DeRaedtD97}
{\sc De~Raedt, L.} {\sc and} {\sc Dehaspe, L.} 1997.
\newblock Clausal {D}iscovery.
\newblock {\em Machine Learning\/}~{\em 26,\/}~2--3, 99--146.

\bibitem[\protect\citeauthoryear{De~Raedt and D\v{z}eroski}{De~Raedt and
  D\v{z}eroski}{1994}]{DeRaedtD94}
{\sc De~Raedt, L.} {\sc and} {\sc D\v{z}eroski, S.} 1994.
\newblock First order jk-clausal theories are {PAC}-learnable.
\newblock {\em Artificial Intelligence\/}~{\em 70}, 375--392.

\bibitem[\protect\citeauthoryear{Dellschaft and Staab}{Dellschaft and
  Staab}{2006}]{DellschaftS06}
{\sc Dellschaft, K.} {\sc and} {\sc Staab, S.} 2006.
\newblock On how to perform a gold standard based evaluation of ontology
  learning.
\newblock In {\em The Semantic Web}, {I.~Cruz}, {S.~Decker}, {D.~Allemang},
  {C.~Preist}, {D.~Schwabe}, {P.~Mika}, {M.~Uschold}, {and} {L.~Aroyo}, Eds.
  Lecture Notes in Computer Science, vol. 4273. Springer, 228--241.

\bibitem[\protect\citeauthoryear{Donini, Lenzerini, Nardi, and Schaerf}{Donini
  et~al\mbox{.}}{1998}]{Donini98}
{\sc Donini, F.}, {\sc Lenzerini, M.}, {\sc Nardi, D.}, {\sc and} {\sc Schaerf,
  A.} 1998.
\newblock $\mathcal{AL}$-log: {I}ntegrating {D}atalog and {D}escription
  {L}ogics.
\newblock {\em Journal of Intelligent Information Systems\/}~{\em 10,\/}~3,
  227--252.

\bibitem[\protect\citeauthoryear{Fisher}{Fisher}{1987}]{Fisher87}
{\sc Fisher, D.} 1987.
\newblock Knowledge acquisition via incremental conceptual clustering.
\newblock {\em Machine Learning\/}~{\em 2,\/}~2, 139--172.

\bibitem[\protect\citeauthoryear{Flach and {Lavra\v{c}}}{Flach and
  {Lavra\v{c}}}{2002}]{FlachL02}
{\sc Flach, P.} {\sc and} {\sc {Lavra\v{c}}, N.} 2002.
\newblock Learning in {C}lausal {L}ogic: {A} {P}erspective on {I}nductive
  {L}ogic {P}rogramming.
\newblock In {\em Computational Logic: Logic Programming and Beyond},
  {A.~Kakas} {and} {F.~Sadri}, Eds. Lecture Notes in Computer Science, vol.
  2407. Springer, 437--471.

\bibitem[\protect\citeauthoryear{Frazier and Page}{Frazier and
  Page}{1993}]{FrazierP93}
{\sc Frazier, M.} {\sc and} {\sc Page, C.} 1993.
\newblock Learnability in inductive logic programming.
\newblock In {\em Proceedings of the 10st National Conference on Artificial
  Intelligence}. The AAAI Press/The MIT Press, 93--98.

\bibitem[\protect\citeauthoryear{Frisch}{Frisch}{1991}]{Frisch91}
{\sc Frisch, A.} 1991.
\newblock The substitutional framework for sorted deduction: Fundamental
  results on hybrid reasoning.
\newblock {\em Artificial Intelligence\/}~{\em 49}, 161--198.

\bibitem[\protect\citeauthoryear{Frisch}{Frisch}{1999}]{Frisch99}
{\sc Frisch, A.} 1999.
\newblock Sorted downward refinement: Building background knowledge into a
  refinement operator for inductive logic programming.
\newblock In {\em {Inductive Logic Programming}}, {S.~D\v{z}eroski} {and}
  {P.~Flach}, Eds. Lecture Notes in Artificial Intelligence, vol. 1634.
  Springer, 104--115.

\bibitem[\protect\citeauthoryear{Frisch and Cohn}{Frisch and
  Cohn}{1991}]{FrischC91}
{\sc Frisch, A.} {\sc and} {\sc Cohn, A.} 1991.
\newblock Thoughts and afterthoughts on the 1988 workshop on principles of
  hybrid reasoning.
\newblock {\em AI Magazine\/}~{\em 11,\/}~5, 84--87.

\bibitem[\protect\citeauthoryear{Gennari, Langley, and Fisher}{Gennari
  et~al\mbox{.}}{1989}]{GennariLF89}
{\sc Gennari, J.}, {\sc Langley, P.}, {\sc and} {\sc Fisher, D.} 1989.
\newblock Models of incremental concept formation.
\newblock {\em Artificial Intelligence\/}~{\em 40,\/}~1-3, 11--61.

\bibitem[\protect\citeauthoryear{G\'{o}mez-P\'{e}rez, Fern\'{a}ndez-L\'{o}pez,
  and Corcho}{G\'{o}mez-P\'{e}rez et~al\mbox{.}}{2004}]{GomezPerez04}
{\sc G\'{o}mez-P\'{e}rez, A.}, {\sc Fern\'{a}ndez-L\'{o}pez, M.}, {\sc and}
  {\sc Corcho, O.} 2004.
\newblock {\em Ontological Engineering}.
\newblock Springer.

\bibitem[\protect\citeauthoryear{Gruber}{Gruber}{1993}]{Gruber93}
{\sc Gruber, T.} 1993.
\newblock A translation approach to portable ontology specifications.
\newblock {\em Knowledge Acquisition\/}~{\em 5}, 199--220.

\bibitem[\protect\citeauthoryear{Halkidi, Batistakis, and Vazirgiannis}{Halkidi
  et~al\mbox{.}}{2001}]{HalkidiBV01}
{\sc Halkidi, M.}, {\sc Batistakis, Y.}, {\sc and} {\sc Vazirgiannis, M.} 2001.
\newblock On clustering validation techniques.
\newblock {\em Journal of Intelligent Information Systems\/}~{\em 17,\/}~2-3,
  107--145.

\bibitem[\protect\citeauthoryear{Han and Fu}{Han and Fu}{1999}]{HanF99}
{\sc Han, J.} {\sc and} {\sc Fu, Y.} 1999.
\newblock Mining multiple-level association rules in large databases.
\newblock {\em IEEE Transactions on Knowledge and Data Engineering\/}~{\em
  11,\/}~5.

\bibitem[\protect\citeauthoryear{Hartigan}{Hartigan}{2001}]{Hartigan01}
{\sc Hartigan, J.} 2001.
\newblock Statistical clustering.
\newblock In {\em International Encyclopedia of the Social and Behavioral
  Sciences}, {N.~Smelser} {and} {P.~Baltes}, Eds. Oxford Press, 15014--15019.

\bibitem[\protect\citeauthoryear{Horrocks, Patel-Schneider, and van
  Harmelen}{Horrocks et~al\mbox{.}}{2003}]{HorrocksP-SvH03}
{\sc Horrocks, I.}, {\sc Patel-Schneider, P.}, {\sc and} {\sc van Harmelen, F.}
  2003.
\newblock From $\mathcal{SHIQ}$ and {RDF} to {OWL}: The making of a web
  ontology language.
\newblock {\em Journal of Web Semantics\/}~{\em 1,\/}~1, 7--26.

\bibitem[\protect\citeauthoryear{Horrocks, Sattler, and Tobies}{Horrocks
  et~al\mbox{.}}{2000}]{HorrocksST00}
{\sc Horrocks, I.}, {\sc Sattler, U.}, {\sc and} {\sc Tobies, S.} 2000.
\newblock Practical reasoning for very expressive description logics.
\newblock {\em Logic Journal of the IGPL\/}~{\em 8,\/}~3, 239--263.

\bibitem[\protect\citeauthoryear{Kietz}{Kietz}{2003}]{Kietz03}
{\sc Kietz, J.} 2003.
\newblock Learnability of description logic programs.
\newblock In {\em Inductive Logic Programming}, {S.~Matwin} {and} {C.~Sammut},
  Eds. Lecture Notes in Artificial Intelligence, vol. 2583. Springer, 117--132.

\bibitem[\protect\citeauthoryear{Langley}{Langley}{1987}]{Langley87}
{\sc Langley, P.} 1987.
\newblock Machine learning and concept formation.
\newblock {\em Machine Learning\/}~{\em 2,\/}~2, 99--102.

\bibitem[\protect\citeauthoryear{Levy and Rousset}{Levy and
  Rousset}{1998}]{LevyR98}
{\sc Levy, A.} {\sc and} {\sc Rousset, M.-C.} 1998.
\newblock Combining {H}orn rules and description logics in {CARIN}.
\newblock {\em Artificial Intelligence\/}~{\em 104}, 165--209.

\bibitem[\protect\citeauthoryear{Lisi}{Lisi}{2006}]{Lisi06-ppswr}
{\sc Lisi, F.} 2006.
\newblock Practice of {I}nductive {R}easoning on the {S}emantic {W}eb: {A}
  {S}ystem for {S}emantic {W}eb {M}ining.
\newblock In {\em Principles and Practice of Semantic Web Reasoning},
  {J.~Alferes}, Ed. Lecture Notes in Computer Science, vol. 4187. Springer,
  242--256.

\bibitem[\protect\citeauthoryear{Lisi and Esposito}{Lisi and
  Esposito}{2004}]{LisiE04-ilp}
{\sc Lisi, F.} {\sc and} {\sc Esposito, F.} 2004.
\newblock Efficient {E}valuation of {C}andidate {H}ypotheses in
  $\mathcal{AL}$-log.
\newblock In {\em Inductive Logic Programming}, {R.~Camacho}, {R.~King}, {and}
  {A.~Srinivasan}, Eds. Lecture Notes in Artificial Intelligence, vol. 3194.
  Springer, 216--233.

\bibitem[\protect\citeauthoryear{Lisi and Esposito}{Lisi and
  Esposito}{2006}]{LisiE06-ecai}
{\sc Lisi, F.} {\sc and} {\sc Esposito, F.} 2006.
\newblock {T}wo {O}rthogonal {B}iases for {C}hoosing the {I}ntensions of
  {E}merging {C}oncepts in {O}ntology {R}efinement.
\newblock In {\em ECAI 2006. Proceedings of the 17th European Conference on
  Artificial Intelligence}, {G.~Brewka}, {S.~Coradeschi}, {A.~Perini}, {and}
  {P.~Traverso}, Eds. IOS Press, Amsterdam, 765--766.

\bibitem[\protect\citeauthoryear{Lisi and Esposito}{Lisi and
  Esposito}{2007}]{LisiE07-ilp}
{\sc Lisi, F.} {\sc and} {\sc Esposito, F.} 2007.
\newblock On the {M}issing {L}ink between {F}requent {P}attern {D}iscovery and
  {C}oncept {F}ormation.
\newblock In {\em Inductive Logic Programming}, {S.~Muggleton}, {R.~Otero},
  {and} {A.~Tamaddoni-Nezhad}, Eds. Lecture Notes in Artificial Intelligence,
  vol. 4455. Springer, 305--319.

\bibitem[\protect\citeauthoryear{Lisi and Malerba}{Lisi and
  Malerba}{2003a}]{LisiM03-aiia}
{\sc Lisi, F.} {\sc and} {\sc Malerba, D.} 2003a.
\newblock Bridging the {G}ap between {H}orn {C}lausal {L}ogic and {D}escription
  {L}ogics in {I}nductive {L}earning.
\newblock In {\em AI*IA 2003: Advances in Artificial Intelligence},
  {A.~Cappelli} {and} {F.~Turini}, Eds. Lecture Notes in Artificial
  Intelligence, vol. 2829. Springer, 49--60.

\bibitem[\protect\citeauthoryear{Lisi and Malerba}{Lisi and
  Malerba}{2003b}]{LisiM03-ilp}
{\sc Lisi, F.} {\sc and} {\sc Malerba, D.} 2003b.
\newblock Ideal {R}efinement of {D}escriptions in $\mathcal{AL}$-log.
\newblock In {\em Inductive Logic Programming}, {T.~Horvath} {and}
  {A.~Yamamoto}, Eds. Lecture Notes in Artificial Intelligence, vol. 2835.
  Springer, 215--232.

\bibitem[\protect\citeauthoryear{Lisi and Malerba}{Lisi and
  Malerba}{2004}]{LisiM04}
{\sc Lisi, F.} {\sc and} {\sc Malerba, D.} 2004.
\newblock Inducing {M}ulti-{L}evel {A}ssociation {R}ules from {M}ultiple
  {R}elations.
\newblock {\em Machine Learning\/}~{\em 55}, 175--210.

\bibitem[\protect\citeauthoryear{Maedche and Staab}{Maedche and
  Staab}{2000}]{MaedcheS00}
{\sc Maedche, A.} {\sc and} {\sc Staab, S.} 2000.
\newblock Discovering {C}onceptual {R}elations from {T}ext.
\newblock In {\em Proceedings of the 14th European Conference on Artificial
  Intelligence}, {W.~Horn}, Ed. IOS Press, 321--325.

\bibitem[\protect\citeauthoryear{Maedche and Staab}{Maedche and
  Staab}{2004}]{MaedcheS04}
{\sc Maedche, A.} {\sc and} {\sc Staab, S.} 2004.
\newblock Ontology {L}earning.
\newblock In {\em Handbook on Ontologies}, {S.~Staab} {and} {R.~Studer}, Eds.
  Springer.

\bibitem[\protect\citeauthoryear{Maedche and Zacharias}{Maedche and
  Zacharias}{2002}]{MaedcheZ02}
{\sc Maedche, A.} {\sc and} {\sc Zacharias, V.} 2002.
\newblock Clustering {O}ntology-{B}ased {M}etadata in the {S}emantic {W}eb.
\newblock In {\em Principles of Data Mining and Knowledge Discovery},
  {T.~Elomaa}, {H.~Mannila}, {and} {H.~Toivonen}, Eds. Lecture Notes in
  Computer Science, vol. 2431. Springer, 348--360.

\bibitem[\protect\citeauthoryear{Mannila and Toivonen}{Mannila and
  Toivonen}{1997}]{Mannila97}
{\sc Mannila, H.} {\sc and} {\sc Toivonen, H.} 1997.
\newblock Levelwise search and borders of theories in knowledge discovery.
\newblock {\em Data Mining and Knowledge Discovery\/}~{\em 1,\/}~3, 241--258.

\bibitem[\protect\citeauthoryear{Michalski and Stepp}{Michalski and
  Stepp}{1983}]{MichalskiS83}
{\sc Michalski, R.} {\sc and} {\sc Stepp, R.} 1983.
\newblock Learning from observation: Conceptual clustering.
\newblock In {\em Machine Learning: an artificial intelligence approach},
  {R.~Michalski}, {J.~Carbonell}, {and} {T.~Mitchell}, Eds. Vol.~I. Morgan
  Kaufmann, San Mateo, CA.

\bibitem[\protect\citeauthoryear{Mitchell}{Mitchell}{1982}]{Mitchell82}
{\sc Mitchell, T.} 1982.
\newblock Generalization as search.
\newblock {\em Artificial Intelligence\/}~{\em 18}, 203--226.

\bibitem[\protect\citeauthoryear{Motik, Sattler, and Studer}{Motik
  et~al\mbox{.}}{2004}]{MotikSS04}
{\sc Motik, B.}, {\sc Sattler, U.}, {\sc and} {\sc Studer, R.} 2004.
\newblock Query {A}nswering for {OWL-DL} with {R}ules.
\newblock In {\em The Semantic Web}, {S.~McIlraith}, {D.~Plexousakis}, {and}
  {F.~van Harmelen}, Eds. Lecture Notes in Computer Science, vol. 3298.
  Springer, 549--563.

\bibitem[\protect\citeauthoryear{N\'{e}dellec, Rouveirol, Ad\'{e}, Bergadano,
  and Tausend}{N\'{e}dellec et~al\mbox{.}}{1996}]{NedellecRABT96}
{\sc N\'{e}dellec, C.}, {\sc Rouveirol, C.}, {\sc Ad\'{e}, H.}, {\sc Bergadano,
  F.}, {\sc and} {\sc Tausend, B.} 1996.
\newblock Declarative bias in {ILP}.
\newblock In {\em Advances in Inductive Logic Programming}, {L.~D. Raedt}, Ed.
  IOS Press, 82--103.

\bibitem[\protect\citeauthoryear{Nienhuys-Cheng and de~Wolf}{Nienhuys-Cheng and
  de~Wolf}{1997}]{Nienhuys97}
{\sc Nienhuys-Cheng, S.} {\sc and} {\sc de~Wolf, R.} 1997.
\newblock {\em Foundations of Inductive Logic Programming}. Lecture Notes in
  Artificial Intelligence, vol. 1228.
\newblock Springer.

\bibitem[\protect\citeauthoryear{Reiter}{Reiter}{1980}]{Reiter80b}
{\sc Reiter, R.} 1980.
\newblock Equality and domain closure in first order databases.
\newblock {\em Journal of ACM\/}~{\em 27}, 235--249.

\bibitem[\protect\citeauthoryear{Rosati}{Rosati}{2005}]{Rosati05}
{\sc Rosati, R.} 2005.
\newblock On the decidability and complexity of integrating ontologies and
  rules.
\newblock {\em Journal of Web Semantics\/}~{\em 3,\/}~1.

\bibitem[\protect\citeauthoryear{Rouveirol and Ventos}{Rouveirol and
  Ventos}{2000}]{RouveirolV2000}
{\sc Rouveirol, C.} {\sc and} {\sc Ventos, V.} 2000.
\newblock Towards {L}earning in {CARIN}-$\mathcal{ALN}$.
\newblock In {\em Inductive Logic Programming}, {J.~Cussens} {and} {A.~Frisch},
  Eds. Lecture Notes in Artificial Intelligence, vol. 1866. Springer, 191--208.

\bibitem[\protect\citeauthoryear{Schmidt-Schauss and Smolka}{Schmidt-Schauss
  and Smolka}{1991}]{Schmidt-Schauss91}
{\sc Schmidt-Schauss, M.} {\sc and} {\sc Smolka, G.} 1991.
\newblock Attributive concept descriptions with complements.
\newblock {\em Artificial Intelligence\/}~{\em 48,\/}~1, 1--26.

\bibitem[\protect\citeauthoryear{Semeraro, Esposito, Malerba, Fanizzi, and
  Ferilli}{Semeraro et~al\mbox{.}}{1998}]{Semeraro98}
{\sc Semeraro, G.}, {\sc Esposito, F.}, {\sc Malerba, D.}, {\sc Fanizzi, N.},
  {\sc and} {\sc Ferilli, S.} 1998.
\newblock A logic framework for the incremental inductive synthesis of
  {D}atalog theories.
\newblock In {\em {Proceedings of 7th International Workshop on Logic Program
  Synthesis and Transformation}}, {N.~Fuchs}, Ed. Lecture Notes in Computer
  Science, vol. 1463. Springer, 300--321.

\bibitem[\protect\citeauthoryear{Staab and Studer}{Staab and
  Studer}{2004}]{StaabS04}
{\sc Staab, S.} {\sc and} {\sc Studer, R.}, Eds. 2004.
\newblock {\em Handbook on Ontologies}.
\newblock International Handbooks on Information Systems. Springer.

\bibitem[\protect\citeauthoryear{Stumme}{Stumme}{2004}]{Stumme04}
{\sc Stumme, G.} 2004.
\newblock Iceberg query lattices for \textsc{Datalog}.
\newblock In {\em Conceptual Structures at Work}, {K.~Wolff}, {H.~Pfeiffer},
  {and} {H.~Delugach}, Eds. Lecture Notes in Artificial Intelligence, vol.
  3127. Springer, 109--125.

\bibitem[\protect\citeauthoryear{Utgoff and Mitchell}{Utgoff and
  Mitchell}{1982}]{UtgoffM82}
{\sc Utgoff, P.} {\sc and} {\sc Mitchell, T.} 1982.
\newblock Acquisition of appropriate bias for inductive concept learning.
\newblock In {\em Proceedings of the 2nd National Conference on Artificial
  Intelligence}. Morgan Kaufmann, Los Altos, CA, 414--418.

\bibitem[\protect\citeauthoryear{Vrain}{Vrain}{1996}]{Vrain96}
{\sc Vrain, C.} 1996.
\newblock Hierarchical conceptual clustering in a first order representation.
\newblock In {\em Foundations of Intelligent Systems}, {Z.~Ras} {and}
  {M.~Michalewicz}, Eds. Lecture Notes in Computer Science, vol. 1079.
  Springer, 643--652.

\bibitem[\protect\citeauthoryear{Xiong, Steinbach, Ruslim, and Kumar}{Xiong
  et~al\mbox{.}}{2005}]{XiongSRK05}
{\sc Xiong, H.}, {\sc Steinbach, M.}, {\sc Ruslim, A.}, {\sc and} {\sc Kumar,
  V.} 2005.
\newblock Characterizing pattern based clustering.
\newblock Technical Report TR 05-015, Dept. of Computer Science and
  Engineering, University of Minnesota, Minneapolis, USA.

\bibitem[\protect\citeauthoryear{Zimmermann and Raedt}{Zimmermann and
  Raedt}{2004}]{ZimmermannDR04}
{\sc Zimmermann, A.} {\sc and} {\sc Raedt, L.~D.} 2004.
\newblock Cluster-grouping: From subgroup discovery to clustering.
\newblock In {\em Machine Learning: ECML 2004}, {J.-F. Boulicaut},
  {F.~Esposito}, {F.~Giannotti}, {and} {D.~Pedreschi}, Eds. Lecture Notes in
  Artificial Intelligence, vol. 3201. Springer, 575--577.

\end{thebibliography}
\end{document}